\documentclass{article}

\PassOptionsToPackage{numbers,compress}{natbib}

\usepackage[preprint]{neurips_2026}

\usepackage[utf8]{inputenc}
\usepackage[T1]{fontenc}
\usepackage{hyperref}
\hypersetup{breaklinks=true,colorlinks=true,linkcolor=black,citecolor=blue,urlcolor=blue}
\usepackage{url}
\usepackage{booktabs}
\usepackage{amsfonts}
\usepackage{nicefrac}
\usepackage{microtype}
\usepackage{xcolor}
\usepackage{graphicx}
\usepackage{subcaption}
\usepackage{xspace}
\usepackage{tikz}
\usepackage{enumitem}
\usepackage{multirow}

\newcommand{\hashtag}[1]{%
  \tikz[baseline=(tag.base)]{
    \node[
      fill=gray!20,
      rounded corners=3pt,
      inner xsep=4pt,
      inner ysep=2pt
    ] (tag) {\strut\##1};
  }%
}
\newcommand{\sys}{VibeServe\xspace}

\title{\sys: Can AI Agents Build \\Bespoke LLM Serving Systems?}

\author{%
  Keisuke Kamahori\thanks{Equal contribution.} \\
  University of Washington
  \And
  Shihang Li\footnotemark[1] \\
  University of Washington
  \And
  Simon Peter \\
  University of Washington
  \And
  Baris Kasikci \\
  University of Washington
}

\begin{document}

\maketitle

\begin{abstract}
For years, we have built LLM serving systems like any other critical infrastructure, with a single general-purpose stack hand-tuned over many engineer-years to support every model and workload.
In this paper, we instead let coding agents build \emph{bespoke} serving systems for different usage scenarios.
We propose \textbf{\sys}, the first agentic loop that generates entire LLM serving stacks end-to-end.
\sys uses an outer loop to plan and track the search over system designs, and an inner loop to implement candidates, check correctness, and measure performance on the target benchmark.
In standard serving settings, \sys matches vLLM and SGLang on Llama-3.1-8B, and delivers $2.3\times$ the goodput of a tuned SGLang for a 397B MoE model on four AMD MI300As.
In five non-standard scenarios, \sys outperforms existing systems by exploiting opportunities that generic systems miss, spanning non-standard model architectures, workload knowledge, and hardware-specific optimizations.
Together, these results show that coding agents can construct and specialize competitive serving systems across heterogeneous deployment targets.
Code is available at \url{https://github.com/uw-syfi/vibesys}.
\end{abstract}

\section{Introduction}
\label{sec:intro}

LLM serving systems are critical software infrastructure for an economy increasingly dependent on generative AI\@.
Open-source stacks such as vLLM~\cite{pagedattention}, SGLang~\cite{sglang}, and TensorRT-LLM~\cite{tensorrt-llm} provide efficient abstractions across a broad range of models and hardware. 
Yet their designs are shaped primarily by mainstream deployments, such as decoder-only Transformers on NVIDIA GPUs serving generic chat workloads.
As a result, emerging model families (e.g., multimodal models or hybrid state-space architectures), along with new hardware accelerators and atypical workloads, often suffer from suboptimal performance or even require substantial new implementation effort \cite{kamahori2026voxserve,yin2026vllm,vellaisamy2025characterizing,jaiswal2025sageserve,gim2025pie,luo2025autellix}.
As the space of model--hardware--workload combinations continues to expand, a one-size-fits-all serving stack increasingly leaves performance on the table for deployments outside the mainstream.

\begin{figure}[t]
\centering
\includegraphics[width=\linewidth]{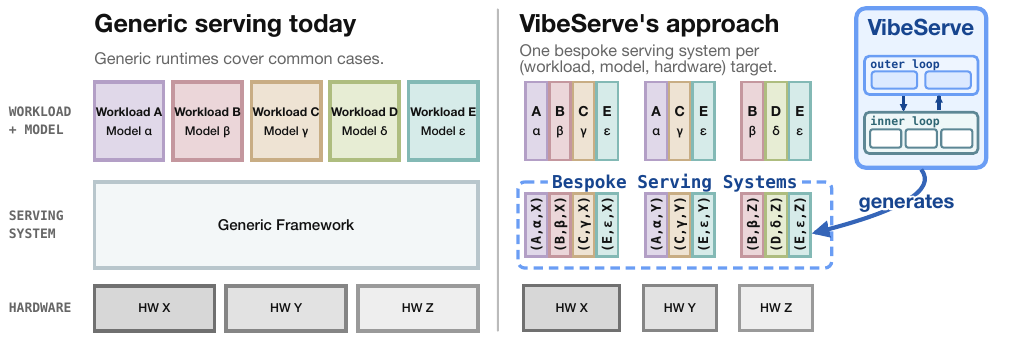}
\caption{%
Motivation for \sys.
General-purpose serving frameworks target common deployments; \sys{} builds systems specialized to each model--hardware--workload target.}
\label{fig:idea}
\end{figure}

In this work, we explore a different point in the design space: \emph{can coding agents build a bespoke serving system for each combination of model, hardware, and workload?}
Per-deployment specialization is a longstanding idea in computer systems~\cite{madhavapeddy2013unikernels,engler1995exokernel,massalin1989threads,bershad1995extensibility,mcnamee2001specialization,madhavapeddy2015jitsu}, but it rarely pays off in practice because per-target engineering cost usually dwarfs the gain.
However, coding agents are changing this calculus. Their demonstrated effectiveness on individual components~\citep{kernelbench,kernelfoundry,alphaevolve,openevolve,glia} and system policies~\citep{schedcp} suggests that per-target specialization could now be feasible at scales where engineering costs were previously prohibitive (Figure~\ref{fig:idea}).

Generating an end-to-end serving system, however, is a long-horizon, multi-component task that existing agentic optimization does not address. Prior systems operate on a much smaller code surface, e.g., a single GPU kernel, an isolated algorithm, or a single policy embedded in an otherwise fixed system~\citep{kernelbench,kernelfoundry,alphaevolve,openevolve,skydiscover2026,glia,engram,schedcp}.
Designing and optimizing an end-to-end system exceeds the context window of any single agent.
The standard recourse, compaction~\citep{anthropic-context-engineering,kangwook-codex-compaction}, can discard details from earlier work and weaken continuity across long tasks~\citep{anthropic-effective-harnesses,wu2026gitcontextcontroller}.
Evolutionary frameworks sidestep this drift via a population of scored programs~\citep{alphaevolve,openevolve,skydiscover2026}, but a scalar score cannot encode the planning state an end-to-end system needs.
Multi-agent loops carry richer state across roles but do not reset agent context windows~\citep{glia,engram}, inheriting limitations from compaction.
Long-horizon coding-agent harnesses sustain state across sessions but produce incorrect systems that underperform state-of-the-art baselines~\citep{anthropic-c-compiler,cursor-scaling-agents}.

We present \textbf{\sys}, a multi-agent system that constructs bespoke LLM serving systems.
To let agents target the open-ended space of model--hardware--workload deployments, \sys exposes two extensible surfaces. The first is a small set of user-provided artifacts (model and reference implementation, accuracy checker, workload benchmark, and target hardware). The second is an Agent Skills library~\cite{agentskills} of serving-systems knowledge distilled from existing engines.
New model families, hardware platforms, and optimization techniques enter as new skill entries, so coverage extends beyond the combinations hand-engineered into existing runtimes.

For each target, \sys factors the work along two axes.
An \emph{outer loop} plans across iterations based on git-recorded optimization history, picking the next optimization and dispatching one concrete task to the inner loop. Its planning state is structured and persistent (e.g., issues, a long-term memory file, the commit history), which is richer than a scalar score and not confined to a single agent's context. This lets the outer loop separate design failures from implementation flaws.
An \emph{inner loop} executes each task through a coding-agent harness. Implementer, Accuracy Judge, and Performance Evaluator agents take turns in fresh contexts, working over a read-only reference implementation and checker.
The outer loop only considers correct implementations. Performance naturally varies as agents explore different design choices, but incorrect candidates cannot derail subsequent rounds.
\S\ref{sec:design} describes the design in detail.

We evaluate \sys across seven scenarios (\S\ref{sec:eval}).
In standard serving settings, \sys matches or beats general-purpose systems. On Llama-3.1-8B-Instruct~\cite{grattafiori2024llama} on H100, it reaches near-parity with vLLM~\cite{pagedattention} and exceeds SGLang~\cite{sglang} by 5\% in throughput. For Qwen3.5-397B on four AMD MI300As, it delivers $2.33\times$ higher goodput than an SGLang that we first ported and tuned, and $1.37\times$ higher than an SGLang optimized by \sys{} under the same time budget.
More importantly, \sys works where generic systems fall short, including non-standard workload patterns (e.g., speculative decoding for code editing with predicted output, workload-aware prompt caching), model architectures (e.g., hybrid attention, complex multimodal models), and hardware backends (e.g., MacBook).
These specialized systems achieve a $5.95\times$ speedup for predicted-output code editing, $3.45\times$ higher throughput for hybrid prompt caching, a $1.69\times$ TTFT speedup for streaming speech recognition, a $2.6\times$ speedup for MacBook JSON decoding, a $6.27\times$ speedup for multimodal inference on MacBook, and a $21.4\%$ latency reduction on H100.

In summary, we contribute the following:
\begin{enumerate}
    \item We argue that per-target bespoke serving is now feasible with long-horizon coding agents.
    \item We build \sys, a multi-agent loop with an outer planner and an inner Implementer/Judge/Evaluator that synthesizes complete serving runtimes.
    \item We demonstrate parity with vLLM and SGLang on standard serving, a $2.33\times$ win over a tuned SGLang on AMD ($1.37\times$ at an equal time budget), and concrete wins across five non-standard scenarios spanning workload, model architecture, and hardware.
\end{enumerate}

\section{Motivation}
\label{sec:background}
\label{sec:portability-tax}

\paragraph{Why LLM serving needs bespoke systems.}
Modern LLM serving stacks~\citep{pagedattention,sglang,tensorrt-llm} achieve strong performance across many models and hardware platforms through various optimization techniques~\citep{orca,pagedattention,flashinfer,dao2022flashattention}.
However, use cases are diversifying rapidly. New model architectures, hardware accelerators, and application interfaces introduce execution structures that challenge runtime abstractions designed for the standard case.
This creates persistent \emph{long-tail} scenarios where a general-purpose stack may work suboptimally, miss optimizations that a bespoke system could implement, or be unable to run the workload at all.
In other words, generic abstractions impose a \emph{portability tax} on non-standard models, hardware, and applications~\citep{engler1995exokernel,mcnamee2001specialization}.

Building bespoke systems can solve this problem.
For example, knowing workload characteristics at design time can enable optimizations that a workload-agnostic runtime cannot safely assume.
RAG-like applications with long shared prefixes can amortize prefill through prompt caching~\citep{gim2024prompt,jin2025ragcache}, while aggressive speculative decoding based on predicted outputs is possible for some applications like code editing~\citep{openai-predicted-outputs,fireworks-cursor,yang2023inference,wang2025efficientedit}.
Similarly, tailoring for a particular model architecture can expose state and execution patterns that fall outside standard decoder-only assumptions.
As an example, hybrid state-space/attention models require cache-management strategies different from those used for decoder-only Transformers~\citep{jamba,nemotronh,marconi,vllm-hybrid-kv,zhang2025jenga}.
Many modern multimodal models also have complex architectures that require significant serving-system effort, such as modality-specific scheduling, memory management, and cross-component execution~\citep{kamahori2026voxserve,yin2026vllm}.
Finally, knowing the target hardware can inform the right runtime design. Apple Silicon, for example, exposes a unified memory model that differs from that of CUDA-centric serving stacks~\citep{mlx,apple-silicon-hpc}.

Exploiting these opportunities is fundamentally a system-level problem.
It requires coordinated decisions across GPU kernels, memory management, request scheduling, and the external interface, and optimizing one component is insufficient if the rest of the runtime still enforces the generic execution contract.
A bespoke system can instead specialize entire layers to the target rather than preserving compatibility with unrelated deployments.

\paragraph{Why bespoke systems are possible now.}
Specialization to remove abstraction overhead is a longstanding idea in computer systems, from extensible operating systems to unikernels~\citep{engler1995exokernel,massalin1989threads,bershad1995extensibility,mcnamee2001specialization,madhavapeddy2013unikernels,madhavapeddy2015jitsu}.
It has been impractical for LLM serving, since building and maintaining a new runtime for every model--hardware--workload combination would normally cost more than it gains.
Recent coding agents suggest a different cost model.
They are increasingly effective at writing software, including real-world bug fixes and performance-critical GPU kernels or scheduling policies~\citep{swebench,xia2024agentless,deng2026swebench,kernelbench,kernelfoundry,alphaevolve,openevolve,glia,schedcp}.
Still, end-to-end system generation remains much more challenging because it requires long-horizon reasoning over a large codebase and coordination across components at all layers~\citep{measuring-ai-ability-to-complete-long-tasks,thai2026sweevo,deng2026swebench,wang2026longhorizontaskmirage,anthropic-effective-harnesses,cursor-scaling-agents,anthropic-c-compiler,engram}.

We argue that LLM serving can be the first domain in which agents successfully generate useful systems end-to-end, since there is a broad need for specialization, the optimization objective is concrete and numeric (e.g., throughput or time-to-first-token latency), and correctness can be checked against a reference implementation. This motivates \sys, which generates a serving system specialized to a given deployment target.

\begin{figure}[t]
    \centering
    \includegraphics[width=\linewidth]{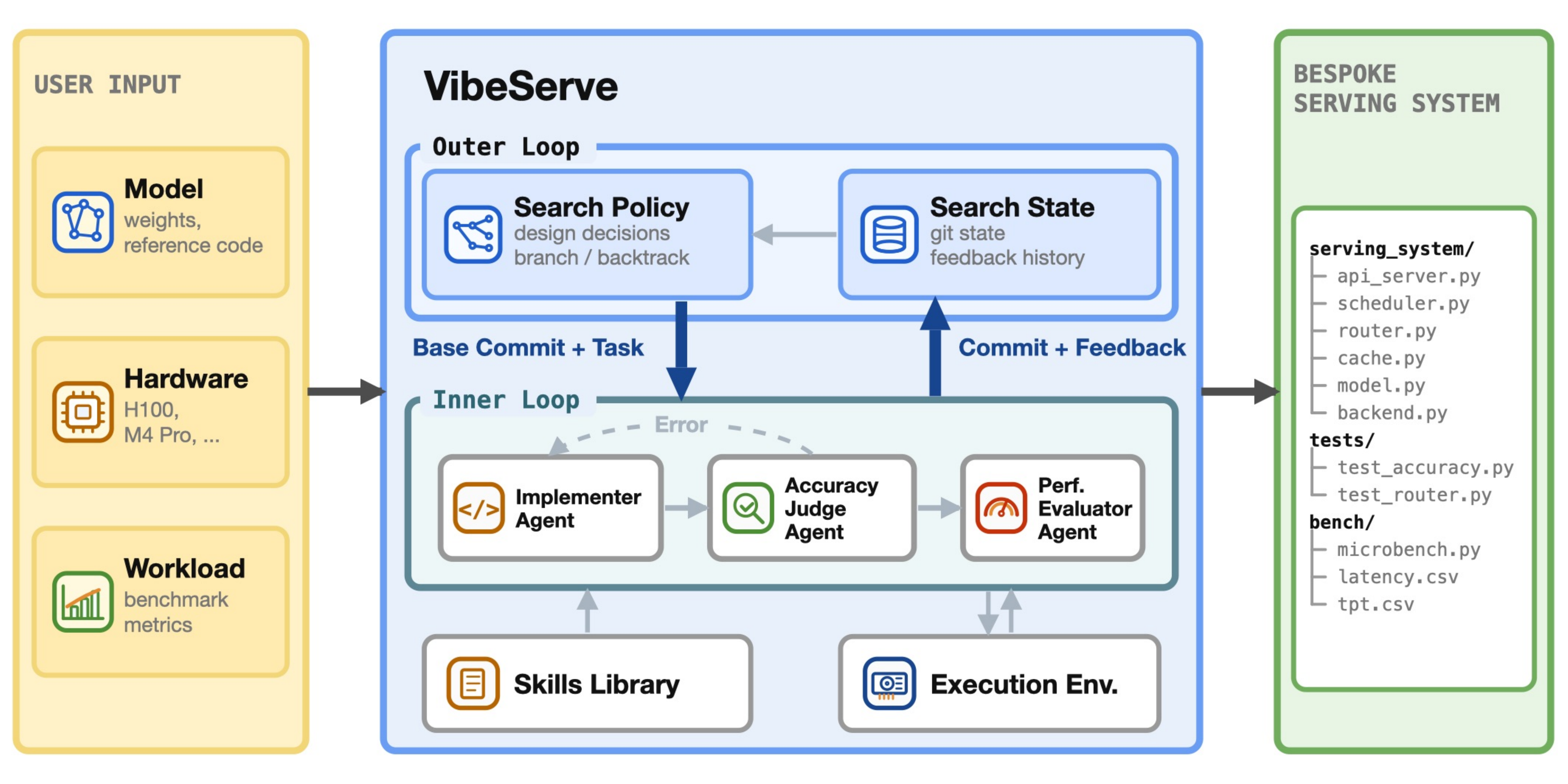}
    \caption{%
    \sys architecture.
    User-provided artifacts define a target deployment.
    The outer loop plans over validated git checkpoints and dispatches a single-round task to the inner loop, where an Implementer, Accuracy Judge, and Performance Evaluator collaborate on a shared workspace using the execution environment and a skills library of serving-systems knowledge.}
    \label{fig:design}
\end{figure}

\section{Design}
\label{sec:design}

Figure~\ref{fig:design} shows \sys's architecture. An outer planning loop and an inner implementation loop iteratively produce an end-to-end serving system from a small set of user-provided artifacts.
The framework is target-agnostic, and specialization enters through three surfaces.
First, per-target \emph{inputs} (\S\ref{sec:design:inputs}) define the model, hardware, and workload.
Second, an \emph{agentic pipeline} (\S\ref{sec:design:agent}) creates and optimizes a bespoke serving system for the target.
Third, an extensible \emph{skills library} (\S\ref{sec:design:skills}) lets agents learn about new model families, hardware platforms, and optimization techniques.

\subsection{Inputs}
\label{sec:design:inputs}

\sys takes a small set of user-provided artifacts that define the target deployment.
First, the user provides the model weights and a reference implementation, such as a Hugging Face Transformers~\citep{wolf2019huggingface} model, which is assumed to be accurate but not efficient.
Second, the user provides an accuracy-checking script that compares a candidate serving system against the reference.
We treat this checker as the source of truth for correctness, mirroring how continuous integration tests gate human-engineered systems.
Fully verifying the semantic accuracy of serving systems is an open problem beyond our scope~\citep{he2025nondeterminism,gond2026llm}.
Third, the user provides a benchmark script that exercises the target workload and emits the metric to optimize, such as latency, throughput, or time-to-first-token.
Finally, the user provides natural-language instructions describing the hardware platform and the expected shape of the deliverable, such as an HTTP API or benchmark harness interface.
Optionally, the user also provides a starting codebase, such as a checkout of an existing serving engine. By default the agents start from an empty repository and synthesize the system end-to-end; with a seed, they specialize the existing engine to the target instead (\S\ref{sec:eval-ablation}).

\subsection{Multi-agent pipeline}
\label{sec:design:agent}

Each candidate runs in an isolated workspace that mounts the user-provided artifacts read-only and exposes the target execution environment (a local or cloud GPU) along with platform-specific profilers (Nsight Systems and the PyTorch profiler on NVIDIA). Agents can only edit the serving-system code they generate, and read-only mounts prevent the Implementer from editing the checker or reference implementation.
\sys wraps existing coding-agent harnesses with three pieces of shared infrastructure: a Model Context Protocol (MCP)~\citep{mcp} server whose schema is defined by the outer-loop policy and through which inner-loop agents return structured information back to the policy, a \emph{skills library} of operational knowledge loaded into the agent context, and an execution environment that issues build, run, and measure calls.

\paragraph{Outer loop.}
The outer loop's search policy is modular, exposing a single per-round operation. It reads prior state, hands the inner loop a starting commit and a task, and receives the resulting commit with the performance metric.
Two shared mechanisms support coordination beyond this contract.
First, every accepted build is a git commit, so any policy can revert cheaply when a later round passes correctness but regresses on the headline metric.
Second, each policy defines its own MCP server schema, whose tools give inner-loop agents a structured channel to return information to the policy during execution.
We implement three policies (evolutionary search~\citep{alphaevolve}, the Ralph loop~\citep{huntley-ralph-loop}, and the issue-tracker policy used in our evaluation).

The \emph{issue-tracker} policy maintains a backlog of structured issues, using the MCP server tool interface to define and enforce the issue schema.
Inner-loop agents file issues through the predefined schema. The outer-loop Orchestrator agent picks the next issue to dispatch at each round, optionally requests a revert to an earlier checkpoint, and updates a long-term memory of optimization directions (a markdown file it reads on entry and edits at the end of each round).
The selected issue, including its acceptance criteria, is the contract handed to the inner loop.
The long-term memory lets the Orchestrator distinguish implementation failures from evidence that a direction is unsuitable for the workload. A failed attempt may signal that the implementation needs debugging or a narrower scope, not that the technique should be discarded.
When issues investigate different hypotheses, the Orchestrator can dispatch several of them to parallel Implementer instances.

\paragraph{Inner loop.}
In the inner loop, Implementer, Accuracy Judge, and Performance Evaluator agents work in sequence, revising the codebase until it passes correctness checks.
This separation keeps implementation, correctness, and performance reasoning in independent contexts. A combined agent can weaken its own correctness criteria to land a hard optimization. The Judge instead inspects diffs and runtime behavior with a fresh context, and the Evaluator runs only after correctness is gated.
Each agent is implemented by a coding-agent harness, such as Codex CLI, Claude Code, or DeepAgents, that can read and edit files, run commands, and return structured results~\citep{openai-codex-cli,anthropic-claude-code,langchain-deepagents}.

The \emph{Implementer} produces and revises the candidate serving system in the workspace.
It receives the task and pass criteria from the outer loop along with pointers to the reference implementation and model weights, and consults the serving-systems skills library (\S\ref{sec:design:skills}).

The \emph{Accuracy Judge} gates correctness in three ways.
It runs the user-provided accuracy checker against the candidate server, verifies any per-round pass criteria from the outer loop (for the issue-tracker policy, the issue's acceptance criteria), and inspects the candidate's source and runtime behavior for reward-hacking patterns such as prompt-keyed completion caches or fast paths that bypass model inference.
On failure, the Judge returns feedback and the inner loop iterates until a build passes or a retry budget is exhausted, at which point the round fails and returns to the Orchestrator.

Once an implementation clears the Judge, the \emph{Performance Evaluator} profiles it and generates performance hints for subsequent rounds.
It starts with end-to-end performance on the user-provided benchmark, then drills down with the platform-specific profilers from the workspace when finer measurements are needed, drawing on the skills library (\S\ref{sec:design:skills}) for profiler-specific guidance.
For targeted investigations, the Evaluator can insert temporary instrumentation around specific code blocks or commit microbenchmarks for repeated measurement. The inner loop then returns the headline metric, trace analysis, hints, and feedback to the outer loop.

\subsection{Skills library}
\label{sec:design:skills}
\sys provides agents with a serving-systems skills library in the Agent Skills format~\citep{agentskills}.
The skills are derived from the source code of mature serving engines and the surrounding research literature, organized along the abstraction layers an engineer works through when building a serving engine (model architectures, serving algorithms, programming frameworks, backend libraries, hardware platforms, reference engines, and tooling).
This lets an agent retrieve focused guidance for a task, such as how continuous batching changes scheduler state, how to use FlashInfer or FlashAttention without reimplementing kernels~\citep{flashinfer,dao2022flashattention}, how MLX differs from PyTorch on Apple Silicon, or where a mechanism lives in vLLM, SGLang, or TensorRT-LLM.

The library is also an extensibility surface.
New model families, hardware platforms, frameworks, backend libraries, or reference engines can be added as new skill entries under the corresponding layer.
Because these axes interact, algorithm skills include compatibility notes linking each technique to supported backends, hardware, and engines (e.g., which paged-KV implementations exist on which hardware).
The library stops at the serving-system boundary. Custom CUDA, Triton, or CUTLASS kernel authoring is delegated to GPU-kernel skills.
Appendix~\ref{app:skills} describes how the library is constructed.

\section{Evaluation}
\label{sec:eval}

Our central question is whether bespoke serving systems generated by \sys are competitive with human-engineered systems and address niche yet important use cases where general-purpose systems fall short.
We evaluate this on seven scenarios spanning the three axes introduced in \S\ref{sec:intro}: workload pattern, model architecture, and hardware.

\subsection{Setup}
All scenarios follow the input interface in \S\ref{sec:design:inputs}. We verify generated systems against the reference implementation and report the workload-relevant metric, such as token throughput, latency, or time-to-first-token (TTFT).
The Implementer, Accuracy Judge, and Performance Evaluator are each instantiated with Codex CLI~\citep{openai-codex-cli} in every scenario except B, which uses Claude Code~\citep{anthropic-claude-code} with Opus 5. The outer loop uses the issue-tracker policy (\S\ref{sec:design:agent}) in every scenario.

We evaluate the following scenarios.

\begin{itemize}[leftmargin=1.5em,itemsep=1pt,topsep=2pt]
    \item \textbf{A: Standard LLM serving.} Llama-3.1-8B-Instruct~\cite{grattafiori2024llama} on H100, a mature setting where existing systems are heavily optimized. We measure generation throughput across arrival rates.

    \item \textbf{B: Distributed MoE serving on AMD.} Qwen3.5-397B-A17B-MXFP4 (512 routed experts, Gated DeltaNet plus attention) on four MI300A accelerators with tensor and expert parallelism, under a multi-turn chat workload that requires preserving KV and DeltaNet state across turns. We measure peak SLO-constrained goodput against an SGLang baseline tuned with \sys{} itself.

    \item \textbf{C: Code editing with predicted outputs.} Qwen3-32B~\cite{yang2025qwen3} on H100 with a predicted-outputs interface~\cite{openai-predicted-outputs}. Code edits overlap heavily with the input~\cite{fireworks-cursor,yang2023inference,wang2025efficientedit}, so speculative decoding from user-provided predictions can skip most decoding steps, a capability absent from standard serving systems. We measure single-batch latency on CodeEditorBench~\cite{guo2024codeeditorbench}. \hashtag{workload}

    \item \textbf{D: Hybrid-architecture prompt caching.} Olmo-Hybrid-7B~\cite{merrill2026olmo} on L4, whose Gated DeltaNet~\cite{yang2024gated} and attention layers make caching hard under limited memory~\cite{marconi}. The workload is RAG-like, with a shared 32k-token prefix. We measure throughput. \hashtag{model} \hashtag{workload}

    \item \textbf{E: Streaming ASR.} Moonshine Streaming medium~\cite{kudlur2026moonshine} on L4, whose sliding-window encoder (unlike Whisper~\cite{whisper}) needs per-stream cache support missing from existing systems. We measure TTFT at concurrency 32 against a vLLM plugin. \hashtag{model} \hashtag{workload}

    \item \textbf{F: Local constrained decoding.} Llama-3.1-8B-Instruct on a MacBook, generating JSON under constrained decoding. Fixed schema spans let a specialized decoder skip generic per-step sampling overhead. We measure single-batch latency on JSONSchemaBench~\cite{jsonschemabench}. \hashtag{workload} \hashtag{hardware}

    \item \textbf{G: Local image generation.} Show-o2~\cite{showo2} on a MacBook and H100, a unified model combining a discrete tokenizer, a diffusion head, and an autoregressive LM that neither vLLM nor vLLM-Omni supports. We measure latency. \hashtag{model} \hashtag{hardware}

\end{itemize}

\subsection{Results}
\label{sec:eval-results}

We present results in scenario order. Iteration-level details are taken from \sys{}'s own logs.

\begin{figure}[t]
  \centering
  \includegraphics[width=0.85\linewidth]{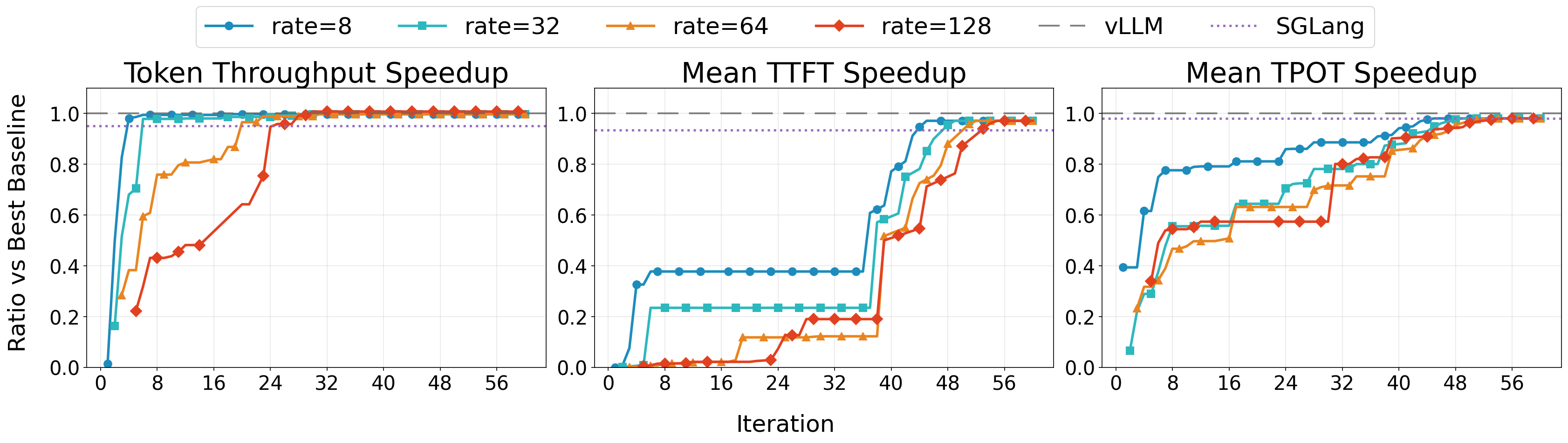}
  \caption{%
On Llama-3.1-8B-Instruct (H100), \sys{} matches vLLM and exceeds SGLang by 5\% (TTFT by 3\%) over 60 agentic-loop iterations.
Panels show \sys{}'s token throughput, mean TTFT, and mean TPOT relative to vLLM ($1.0$ is parity, higher is better), at four request rates.}
  \label{fig:standard-convergence}
\end{figure}

\paragraph{Scenario A: parity on a heavily optimized setting.}
\label{sec:eval-standard}
Figure~\ref{fig:standard-convergence} traces 60 \sys{} iterations on Llama-3.1-8B-Instruct (H100). The generated system reaches vLLM parity on token throughput and TPOT at all four request rates and lands within 5\% on TTFT. It exceeds SGLang by 5\% on throughput and 3\% on TTFT. \sys{} pursued throughput first, reaching parity by iteration~30 with latency roughly flat, then shifted to latency, with TTFT and TPOT improving sharply over iterations~30--60. \sys{} introduced each higher rate after plateauing, escalating to $128$~req/s.

\paragraph{Scenario B: distributed MoE serving on AMD.}
\label{sec:eval-amd-dist}
\begin{figure}[t]
  \centering
  \includegraphics[width=0.95\linewidth]{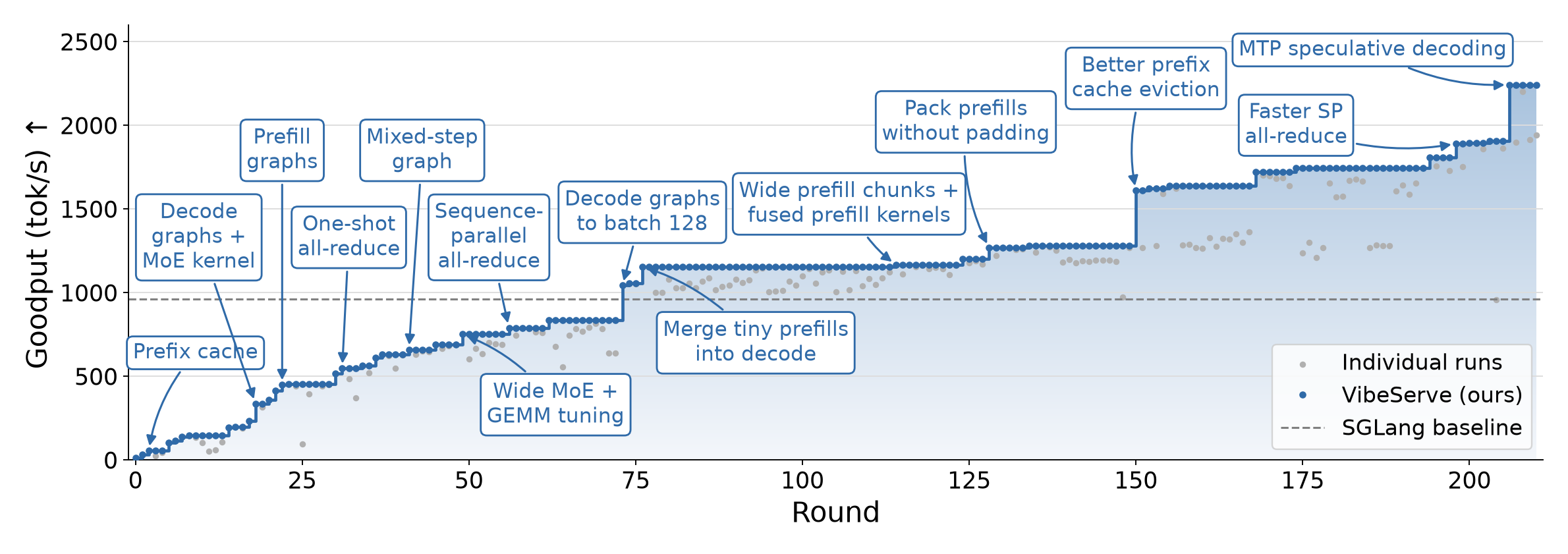}
  \caption{Scenario B. Best goodput across the load ramp at each round, on 4$\times$ AMD MI300A. The dashed line is the tuned SGLang baseline ($961$~tok/s). Gray dots are individual runs.}
  \label{fig:amd-dist-throughput}
\end{figure}
Figure~\ref{fig:amd-dist-throughput} shows peak goodput on Qwen3.5-397B-A17B-MXFP4 across four AMD MI300A accelerators on the multi-turn chat workload detailed in \S\ref{app:eval-scenarios}. Starting from a simple PyTorch implementation at $12.8$~tok/s, \sys{} adds prefix caching that reuses both attention KV and DeltaNet state, decode and prefill graph capture with a fused MoE kernel, a series of communication optimizations (one-shot and sequence-parallel all-reduce), decode graphs up to batch size 128, folding of small turn-opening prefills into decode steps, wide and unpadded prefill packing, better prefix-cache eviction, and finally speculative decoding with the model's multi-token-prediction (MTP) head. The result is $2{,}242$~tok/s, which is $2.33\times$ the tuned SGLang baseline ($961$~tok/s) and $\sim$$175\times$ the starting point. Stock SGLang does not run this model on MI300A, so the baseline is a \sys{}-modified version that we first made functional and then tuned (\S\ref{app:eval-scenarios}), with minimal changes to make it work. Both systems use MTP with $k{=}3$. We ramp the load for each system and report the best goodput. SGLang peaks at 48 concurrent chat sessions, while our engine peaks at 96. Under an equal 105-hour budget, the \sys-optimized bespoke engine reaches $1.37\times$ the goodput of a \sys-optimized SGLang (Appendix~\ref{app:extend-sglang-b}).

\begin{figure}[t]
  \centering
  \begin{subfigure}[t]{0.33\linewidth}
    \centering
    \includegraphics[width=\linewidth]{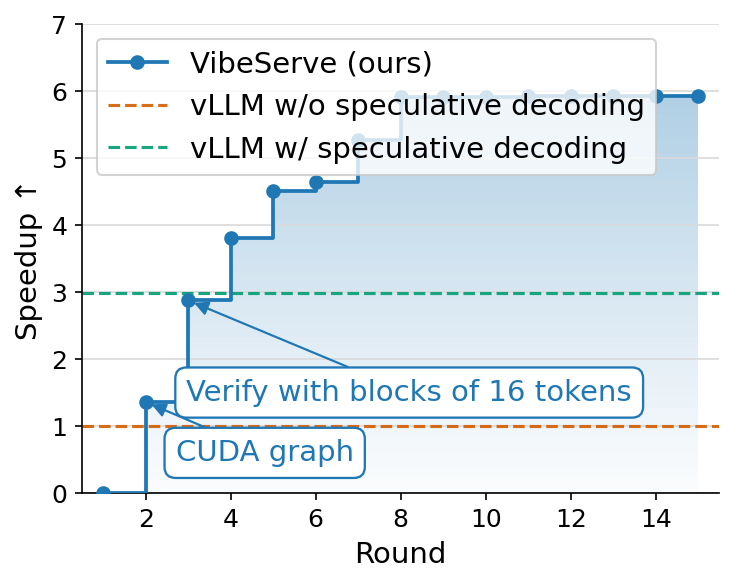}
    \caption{Scenario C.}
    \label{fig:qwen-code-edit-latency}
  \end{subfigure}\hfill
  \begin{subfigure}[t]{0.33\linewidth}
    \centering
    \includegraphics[width=\linewidth]{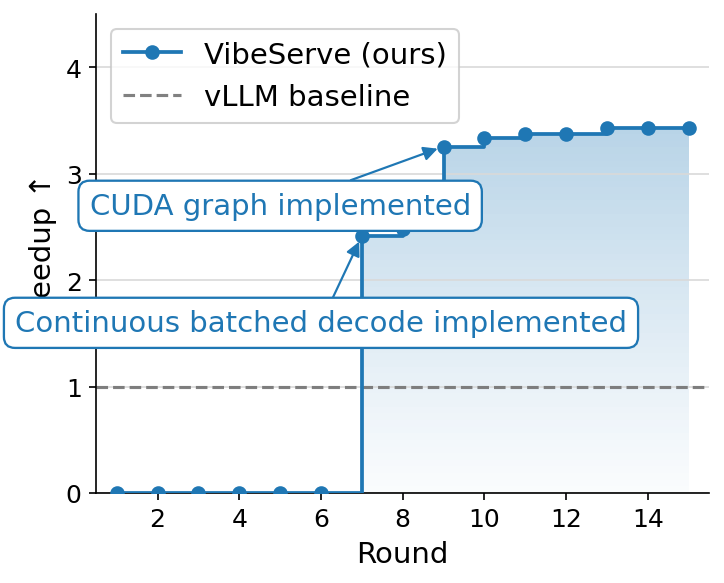}
    \caption{Scenario D.}
    \label{fig:olmo-hybrid-latency}
  \end{subfigure}\hfill
  \begin{subfigure}[t]{0.33\linewidth}
    \centering
    \includegraphics[width=\linewidth]{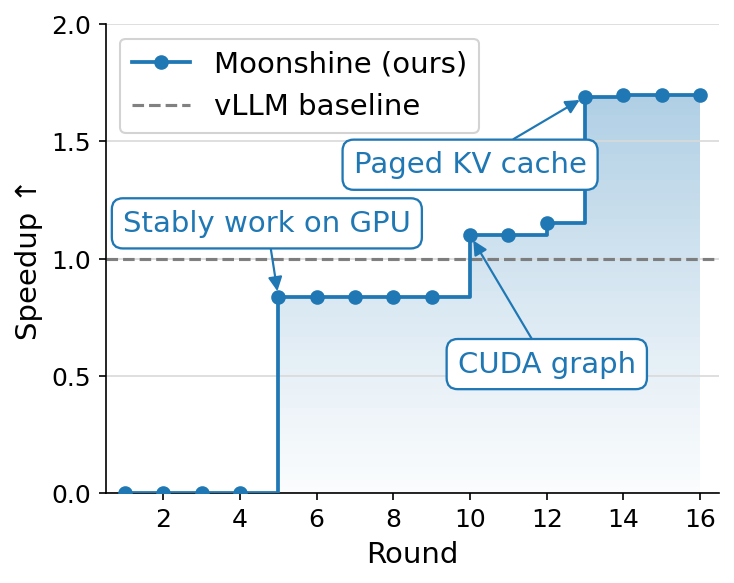}
    \caption{Scenario E.}
    \label{fig:moonshine-streaming-latency}
  \end{subfigure}\hfill
  \caption{%
Workload- and model-specific scenarios: speedup over a baseline across \sys{} iterations (dashed line is parity; higher is better). (a)~Qwen3-32B vs.\ vLLM without/with draft-model speculative decoding. (b)~Olmo-Hybrid-7B vs.\ vLLM. (c)~Moonshine TTFT vs.\ a vLLM plugin.}
  \label{fig:workload-model-results}
\end{figure}

\paragraph{Scenario C: predicted-output speculative decoding.}
\label{sec:eval-qwen-code-edit}
Figure~\ref{fig:qwen-code-edit-latency} traces 15 iterations on Qwen3-32B/CodeEditorBench against two vLLM baselines: vanilla autoregressive ($1.0\times$) and draft-model speculative decoding ($\approx 3.0\times$, dashed).
After an early CUDA-graph pass, iteration~3 introduces the predicted-output verifier. It proposes tokens from the user-supplied prediction in 16-token blocks and verifies each block in a single target-model forward. This reaches $2.9\times$, already on par with vLLM's draft-model speculative decoder at zero draft-model compute.
Block sizing and acceptance bookkeeping then push this to $5.95\times$ by iteration~14, $2.0\times$ over vLLM-with-spec-dec.

\paragraph{Scenario D: hybrid-architecture prompt caching.}
\label{sec:eval-olmo-hybrid}
Figure~\ref{fig:olmo-hybrid-latency} shows token-generation throughput vs.\ vLLM on Olmo-Hybrid-7B (L4) over 15 iterations.
The first six rounds fail accuracy gates while \sys{} wires up the dual cache, with attention KV blocks plus per-DeltaNet recurrent-state snapshots at the prefix boundary.
Continuous batched decode against the shared state, then CUDA-graph capture, bring the system to a $3.45\times$ plateau.
vLLM's GDN prefix caching cannot share DeltaNet state at this workload's shared/unique boundary (\S\ref{app:eval-scenarios}), so the baseline recomputes the 32k prefix per request.

\paragraph{Scenario E: streaming ASR.}
\label{sec:eval-moonshine}
Figure~\ref{fig:moonshine-streaming-latency} shows TTFT speedup over a vLLM-Moonshine plugin at concurrency~32 on L4 over 16 iterations.
\sys{} first reaches a working but sub-baseline configuration ($0.84\times$) by aligning the per-stream encoder cache with Moonshine's sliding-window attention, then adds CUDA-graph capture and a paged KV cache for per-stream encoder state to reach $1.69\times$.
The improvement comes from giving the encoder layer first-class per-stream cache management, which the plugin path does not expose.

\begin{figure}[t]
  \centering
  \begin{subfigure}[t]{0.31\linewidth}
    \centering
    \includegraphics[width=\linewidth]{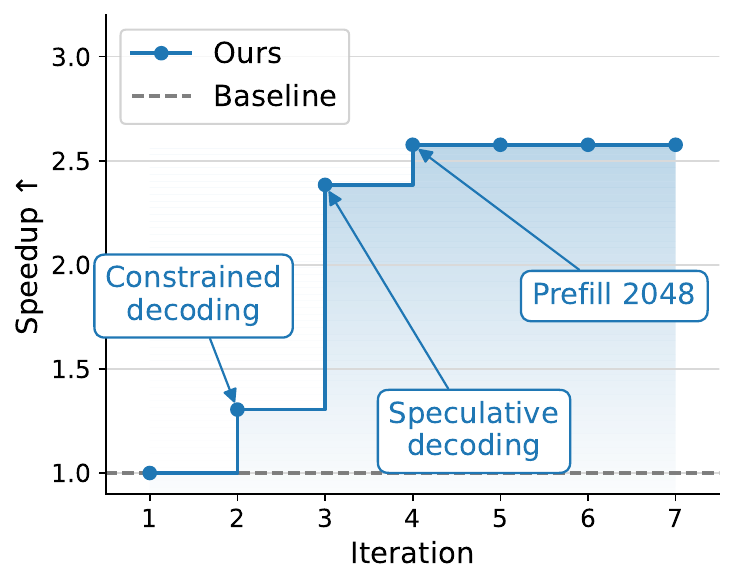}
    \caption{Scenario F.}
    \label{fig:llama-spec-mbp}
  \end{subfigure}\hfill
  \begin{subfigure}[t]{0.31\linewidth}
    \centering
    \includegraphics[width=\linewidth]{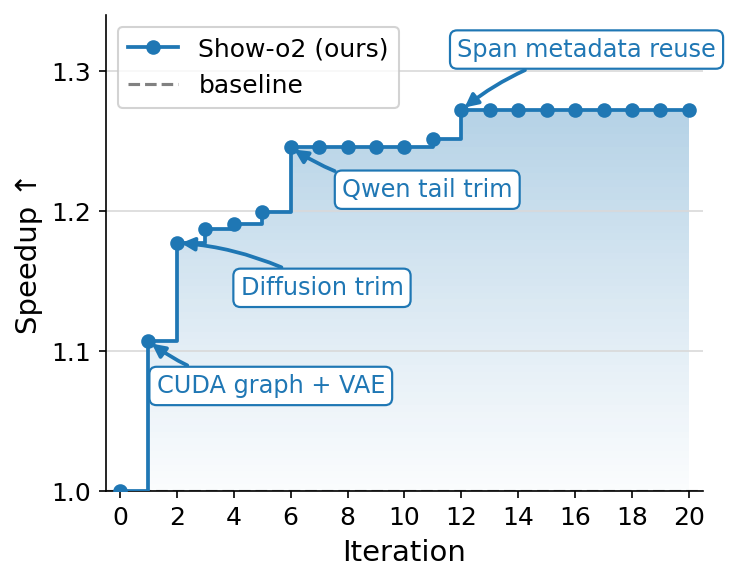}
    \caption{Scenario G, H100.}
    \label{fig:show-o2-h100-latency}
  \end{subfigure}\hfill
  \begin{subfigure}[t]{0.31\linewidth}
    \centering
    \includegraphics[width=\linewidth]{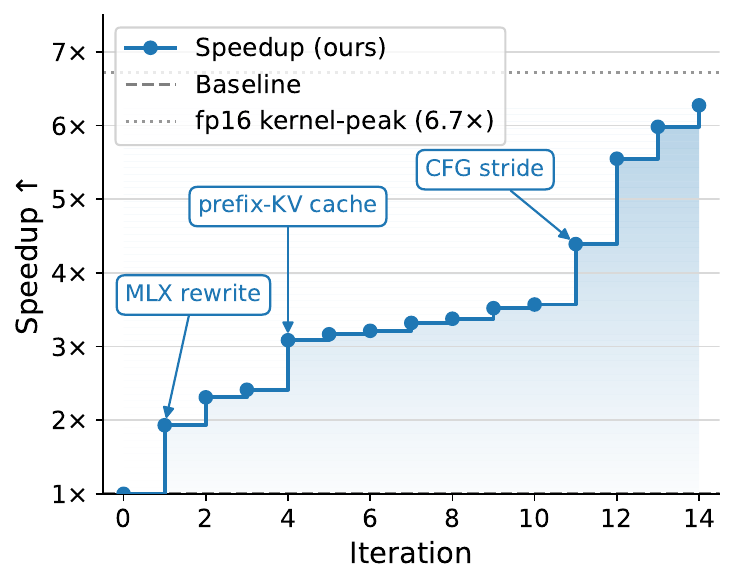}
    \caption{Scenario G, MacBook.}
    \label{fig:show-o2-mbp-latency}
  \end{subfigure}\hfill
  \caption{%
Hardware- and workload-specific scenarios with speedup over a baseline across \sys{} iterations. The dashed line at $1.0$ is parity, and higher is better. (a)~Llama-3.1-8B-Instruct JSON decoding on a MacBook (Apple M3 Pro, 36\,GB). (b)~Show-o2 1.5B-HQ $432{\times}432$ text-to-image on H100. (c)~Show-o2 on the MacBook, where the dotted line is the fp16 kernel-peak ceiling ($6.7\times$).}
  \label{fig:hardware-results}
\end{figure}

\paragraph{Scenario F: constrained JSON decoding on a MacBook.}
\label{sec:eval-spec-cons-mbp}
Figure~\ref{fig:llama-spec-mbp} traces the trajectory from a $22.1$\,s vanilla autoregressive baseline.
\sys{} first adds XGrammar-based constrained decoding~\citep{li2026xgrammar2efficientdynamicstructured} ($16.9$\,s), then layers speculative decoding with a Llama-3.2-1B-Instruct-4bit draft against the 8B-8bit target at $K{=}4$, reaching $9.3$\,s. A larger 3B-4bit draft was slower, since the 1B's lower per-step cost outweighed its lower acceptance rate.
Raising \texttt{mlx\_lm}'s \texttt{prefill\_step\_size} from 512 to 2048 lets \sys{} prefill our $\sim$1300-token prompts in one chunk, yielding $8.6$\,s ($2.6\times$). K/V quantization, alternative $K$, and \texttt{mx.compile} did not help.

\paragraph{Scenario G: Show-o2 on H100 and MacBook.}
\label{sec:eval-show-o2}
On H100 (Figure~\ref{fig:show-o2-h100-latency}), p50 latency falls from $873$\,ms to $687$\,ms ($21.4\%$) over 20 iterations.
Gains are front-loaded in the first dozen rounds (CUDA-graph replay/prewarm, trimming inactive diffusion tokens, restricting AdaLN to the active image span, and a Qwen tail trim). Further passes then map the limits, as changes that regress quality or alter outputs are rejected and Qwen prefix reuse yields no gain (\S\ref{app:eval-scenarios}).

\label{sec:eval-show-o2-mac}
On MacBook (Figure~\ref{fig:show-o2-mbp-latency}), \sys{} first ports the Qwen2.5-1.5B body and 10-block diffusion head to MLX and elides a redundant SigLIP \texttt{und\_trans} pass on noisy latents ($2.4\times$). Cross-step redundancy then dominates, and prefix-KV caches on the body and head, plus a prefill trim to \texttt{[0,\,image\_end)}, bring warm latency to $3.5\times$, with the body at $\sim$92\% of the fp16 compute peak. A classifier-free-guidance (CFG) stride at $K{=}16$ skips the unconditional branch on $K{-}1$ of every $K$ steps and reuses the cached \texttt{v\_uncond}. This reaches $15.54$\,s ($6.27\times$ over PyTorch-MPS), within $\sim$7\% of a $14.5$\,s fp16 kernel-perfect floor.

\subsection{Ablations}
\label{sec:eval-ablation}

We ablate the design choices of \S\ref{sec:design}. The pipeline ablations run on Scenario A (Llama-3.1-8B, H100, Codex GPT-5.5 Medium), which isolates pipeline behavior from target-specific optimizations. None reaches a competitive serving system, whereas the full pipeline matches vLLM ($\approx$10k tok/s) after 60 rounds / 25.0\,h (Table~\ref{tab:loop-breakdown}). \S\ref{app:extend-vllm} gives full details.

\paragraph{Starting point: from scratch vs.\ an existing codebase.}
\sys's inner loop can also start from an existing codebase. We seed Scenarios D and E from vLLM v0.26.0 and Scenario G from vLLM-Omni v0.24.1 and compare against the from-scratch runs (\S\ref{app:extend-vllm}). Seeding helps only when the seed's abstractions already fit the target. vLLM already supports Olmo-Hybrid, so the Scenario~D agent only adds workload-specific prefix caching and matches from-scratch performance by round~4, roughly twice as fast in wall-clock time. For Moonshine (E) and Show-o2 (G), where the seed lacks support, the agents implement outside the seed's abstractions rather than risk invasive edits.

\paragraph{Agentic pipeline, skills library, and outer loop.}
\emph{Single-agent.} Collapsing the Implementer, Judge, and Performance Evaluator into one long-running agent reaches only 617~tok/s over 35~hours. Without a fresh-context Judge, it repeatedly skips its own accuracy and benchmark gates and gets stuck in local optimization rather than attempting larger architectural changes. This supports factoring implementation and validation into separate agents (\S\ref{sec:design:agent}).

\emph{No-skill.} Removing the skills library caps throughput at 1{,}100~tok/s after $\sim$35 rounds. Agents rediscover standard techniques from the codebase alone but, without pointers from skill files, focus on minor parameter tuning rather than the larger architectural changes that close the remaining gap. This shows the skills library (\S\ref{sec:design:skills}) supplies directions that experimentation alone does not surface.

\emph{Outer-loop $\to$ OpenEvolve.} Replacing the issue-tracker orchestrator with OpenEvolve~\citep{openevolve}, an AlphaEvolve-style~\citep{alphaevolve} evolutionary search over scored populations, reaches only 537~tok/s after 30 rounds. It mutates a fixed, user-marked code region that does not scale past 1k~LoC, and its population-based search lacks the issue-tracker policy's long-term memory (\S\ref{sec:design:agent}) for planning across rounds.

\section{Related Work}
\label{sec:related}

Agentic optimization systems use a few search paradigms, none applied to greenfield end-to-end system synthesis.
\emph{Evolutionary search} selects among agent-generated candidates by measured performance~\citep{alphaevolve,openevolve,skydiscover2026,kernelfoundry,evoengineer}. \emph{Multi-agent iteration} has agents hypothesize, experiment, and refine across rounds within a single context window~\citep{glia,engram}. \emph{Autoresearch}~\citep{karpathy-autoresearch} uses one long-running agent that tracks candidates across git branches.
All three target a bounded code scope (e.g., a marked region) or rely on a scalar score or a single conversation that cannot encode the bottleneck information driving an end-to-end system's next step.
\sys is the first agentic system to design a multi-component serving system end-to-end.

\sys sits within a broader literature on long-horizon coding agents~\citep{measuring-ai-ability-to-complete-long-tasks,thai2026sweevo,deng2026swebench,anthropic-effective-harnesses,huntley-ralph-loop,wu2026gitcontextcontroller,anderson2025self}.
Tasks that exceed a context window are typically handled by compaction~\citep{anthropic-context-engineering,kangwook-codex-compaction}, whose lossy summarization causes drift in performance and correctness.
Industrial prototypes from Cursor and Anthropic build end-to-end systems by handing work between fresh agent sessions through task abstractions over shared repository state~\citep{cursor-scaling-agents,anthropic-c-compiler}, but stop short of optimizing performance.
Building on these ideas, \sys targets \emph{performant} code by giving agents direct profiler access, folding performance analysis into every implementation change through role-based agents, and packaging platform, optimization, and profiling knowledge as skills (Appendix~\ref{app:related}).

\section{Conclusion}
\label{sec:conclusion}

\sys demonstrates that coding agents can construct and specialize competitive serving systems across heterogeneous deployment targets. \sys matches vLLM and beats a tuned SGLang in standard settings, and yields concrete wins across five non-standard scenarios, two of which cannot run on any generic stack. Existing engines provide an effective starting point when their abstractions fit the target (\S\ref{sec:eval-ablation}); synthesis provides a complementary path when the target requires substantial changes to execution, state management, or scheduling.

\sys relies on a user-supplied correctness checker and a non-trivial per-target compute budget (\S\ref{app:eval-scenarios}), was evaluated only on a single node, and leaves the maintainability of generated systems unmeasured. Looking ahead, as coding agents improve, we expect agent-driven specialization, whether it extends an existing engine or synthesizes a new one, to pay off~\cite{madhavapeddy2013unikernels,engler1995exokernel,massalin1989threads} in more domains where generic abstractions cost performance.

\section*{AI Use Statement}
We used Codex CLI and Claude Code to aid and polish writing, retrieve and
discover related work, and support research ideation and execution. In the
reported experiments, these coding agents generated and revised serving-system
implementations, ran accuracy and performance checks, and analyzed measurements
to guide subsequent optimization. We evaluated the generated systems using the
accuracy checkers and benchmarks described in Sections~\ref{sec:design} and
\ref{sec:eval}. The authors take
responsibility for the final text, citations, experiments, and claims.

{
\small

\bibliographystyle{plainnat}
\bibliography{references}

}

\appendix
\section{Detailed Evaluation Scenarios}
\label{app:eval-scenarios}

This appendix provides additional details for each evaluation scenario. Across scenarios, the agentic loop receives model weights, a HuggingFace Transformers reference implementation~\cite{wolf2019huggingface}, accuracy-checking scripts, performance-evaluation scripts, and natural-language instructions. The generated system is evaluated against the reference implementation for correctness and against one or more baseline serving systems for performance.

\paragraph{Scenario A: Standard LLM serving on H100.}
\emph{Architecture.} Llama-3.1-8B-Instruct~\cite{grattafiori2024llama} is a dense decoder-only Transformer with grouped-query attention (32 query heads sharing 8 key/value heads), RoPE positional encodings, SwiGLU MLPs, and a 128k-token context window. This configuration is the design center of every modern serving stack: dense decoder-only inference on data-center GPUs is precisely what vLLM, SGLang, and TensorRT-LLM are tuned for, and the standard optimization stack is by now well-known --- paged KV cache~\cite{pagedattention}, continuous batching~\cite{orca}, CUDA graphs, FlashAttention/FlashInfer kernels~\cite{dao2022flashattention,flashinfer}, and operator fusion.

\emph{Workload and metric.} An open-loop synthetic load generator drives the system at four request rates (8, 32, 64, 128 req/s). Request arrivals follow a Poisson process (exponential inter-arrival times), and requests are launched independently of completion. Each rate is run for 60 seconds with seed 42. Prompts are sampled uniformly from a predefined prompt pool, and output length is capped at \texttt{max\_tokens=128}. Most requests emit exactly 128 chunks, and generation uses temperature 0. We report token-generation throughput, mean TTFT, and mean TPOT relative to vLLM. Greedy decoding outputs are checked against the Hugging Face Transformers reference.

\paragraph{Scenario B: Distributed MoE serving on AMD MI300A.}
\emph{Architecture.} Qwen3.5-397B-A17B-MXFP4 is a sparse mixture-of-experts model with approximately 397B total parameters, approximately 17B active parameters per token, and 512 routed experts in MXFP4 precision. It interleaves Gated DeltaNet linear-attention layers with recurrent state, full-attention layers with paged KV state, and shared/routed MoE computation, so prefix reuse must preserve both conventional attention KV and DeltaNet recurrent state at once. The engine runs on one node of 4$\times$ AMD MI300A (ROCm, gfx942, unified host/accelerator memory): dense projections, attention, and DeltaNet are tensor-sharded across the four ranks, and MoE experts are partitioned across the same ranks with partial outputs combined collectively.

\emph{Workload and metric.} A deterministic multi-turn chat benchmark. Concurrent sessions are ramped from 1 to 192 in a load-ramp benchmark, with 3--6 turns per session, 80--300 generated tokens per turn, 1--8\,s of think time between turns, the full accumulated conversation resent on every turn, and greedy decoding at fixed completion lengths. A response fails the run if it is dropped, errored, empty, or truncated, or if it misses the SLO of p95 TPOT $\leq 250$\,ms and p95 TTFT $\leq 10$\,s. We report peak goodput in tokens/s, the token throughput of SLO-compliant responses, taking the best value across the load ramp for each system. The comparison is against an optimized SGLang v0.5.18 baseline using MTP speculative decoding ($k{=}3$). SGLang peaks at 48 concurrent sessions, while our engine peaks at 96, where we report $2{,}242$~tok/s. Every accepted change must also pass a numerics gate, with at least 1{,}074 of 1{,}088 teacher-forced top-1 tokens matching the reference and 5 of 5 greedy reference completions. Because node-to-node variation is roughly $\pm10\%$, gains are measured as same-node paired runs. Performance evaluation uses a 20\% held-out set.

\emph{Optimization opportunity.} Most of the gain from $12.8$ to $2{,}242$~tok/s comes from a long campaign of general serving-engine engineering, including HIP-level decode and prefill graph capture, fused MoE and attention kernels, chunked DeltaNet recurrence, split-K attention, per-shape GEMM tuning, and an overlap scheduler. Because the model is tensor- and expert-sharded across four ranks, cross-GPU communication is a first-order cost here in a way it is not in the other scenarios, and a substantial share of the campaign's gains are communication-focused, including a custom all-reduce collective, fusing the decode-path all-reduce with the residual add and RMSNorm into a single collective, wide HIP MoE kernels, rank-agreed validation, and a vocab-parallel LM head. Later, workload-specific refinements build on this tuned base. Because every turn resends the entire conversation, the engine reuses both the attention KV cache and the DeltaNet recurrent state from prior turns rather than recomputing them, and folds the turn-opening chat-template suffix (roughly five tokens) into decode steps already scheduled for the active batch instead of issuing it as a separate small prefill call ($9.6\%$ on top of an already-tuned system). A DeltaNet state snapshot costs about 47\,MiB per conversation point, so the prefix cache holds far fewer conversations than it would for a pure-attention model, and improving its eviction policy gives a further large jump. The final step is speculative decoding with the model's built-in MTP head. Making it correct required exact DeltaNet recurrence and causal convolution over the drafted tokens, plus rollback of recurrent state on rejected drafts that stays safe under graph capture.

\emph{Why stock SGLang does not run.} SGLang v0.5.18 does not serve this model on MI300A as released. The main blocker is the expert path. Its MXFP4 MoE relies on AITER kernels, and on gfx942 these abort with an unsupported fp4x2 output type. The CK a4w4 kernels are built for gfx950 only, the FlyDSL a4w4 path fails, and CK's MXFP4 grouped GEMM needs the gfx950 scaled MFMA instruction. No shipped kernel can consume MXFP4 weights on this part. Several smaller problems also block a usable launch. The shipped AITER GEMM tuning table is keyed to gfx950 with 256 CUs while MI300A reports gfx942 with 228, so no row matches and dense projections fall back to untuned kernels. Checkpoint loading takes about an hour because per-tensor expert materialization is serialized, far above the 22\,s storage floor. Rank initialization crashes when setting GPU process affinity under a Slurm-restricted cpuset. The hybrid-model cache sizing fails with a negative Mamba cache size when free memory is uneven across ranks. AITER also builds some attention kernel variants lazily, which stalls the first real request by about 105\,s. \sys{} first replaced the expert path with a Triton w4a16 fused-MoE kernel that dequantizes MXFP4 in-kernel, which brought the server up at about 28\,tok/s on a single stream. It also threaded the expert weight loader, which cut load time to about 4 minutes, and fixed the remaining launch failures. Profiling showed the Triton kernel at roughly $26\times$ above its byte bound and limited by dequantization and latency rather than bandwidth, and retuning its tiling did not help.

\emph{Tuning baseline SGLang.} We ran \sys{} on SGLang itself before comparing. It replaced the gfx942 Triton MoE fallback with a custom fused HIP MXFP4 MoE kernel, added a skinny BF16 GEMM path for small-batch decode, enabled the model's NEXTN/MTP speculative-decoding head, generated PyTorch TunableOp GEMM tables for every captured draft/verify/decode shape, padded and re-bucketed prefill GEMMs, and made other changes. Once SGLang ran, this tuning improved its goodput by a further 42\%. The reported $2.33\times$ gain is measured against this \sys{}-tuned version.

\emph{Why the tuned SGLang still falls short.} The remaining gap comes from changes that SGLang's serving abstractions cannot express as configuration or kernel substitution. Jointly placing DeltaNet recurrent state and MoE expert shards across the same four ranks under MI300A's unified host/accelerator memory --- where snapshot pools, KV pools, captured-graph allocations, and process RSS all compete for the same physical capacity --- requires a memory layout that SGLang's parallelism and cache-allocation interfaces do not expose as tuning knobs. Folding the turn-opening chat-template suffix into an already-scheduled decode step similarly requires restructuring the request lifecycle itself, which means invasive SGLang changes. The system generated by \sys{} implements both directly.

\paragraph{Scenario C: Code editing with predicted outputs.}
\emph{Architecture and workload.} 
Qwen3-32B~\cite{yang2025qwen3} is a dense decoder-only Transformer (with Q/K-norm and grouped-query attention) served on an NVIDIA H100. The workload is code editing under OpenAI's predicted-outputs interface~\cite{openai-predicted-outputs}: each request carries both an instruction and a string of \emph{predicted output tokens} representing the most likely answer. This prediction is naturally available for code editing tasks, since the pre-edit file is typically a near-prediction of the post-edit file, and prior work and deployed systems show that the overlap is large in practice~\cite{fireworks-cursor,yang2023inference,wang2025efficientedit}. We report single-batch latency on CodeEditorBench~\cite{guo2024codeeditorbench}.

\emph{Optimization opportunity.} 
The predicted-outputs interface is a degenerate case of speculative decoding in which the draft is the user-supplied prediction at zero draft-model cost. The serving system feeds a window of $K$ predicted tokens through the target model in a single forward pass and commits the longest prefix whose argmax matches the prediction. On a mismatch, it falls back to ordinary autoregressive decoding for one token and then resumes from the prediction. With high overlap, latency drops by nearly a factor of $K$ with no additional compute.

\emph{Why generic systems cannot exploit this.} 
Standard systems like vLLM or SGLang support speculative decoding, but do not support the predicted-outputs interface. Predicted outputs are a different request type. The engine needs a per-request token stream, a verifier loop that consumes from that stream until divergence, and well-defined fallback semantics. Adding this to vLLM means non-trivial changes to the scheduler, sequence-group state, and sampler. A bespoke system can build the request lifecycle directly around the predicted-outputs API.

\paragraph{Scenario D: Prompt caching for a hybrid architecture.}
\emph{Architecture.} Olmo-Hybrid-7B~\cite{merrill2026olmo} interleaves Gated DeltaNet layers~\cite{yang2024gated} with standard self-attention layers. Gated DeltaNet is a linear-attention/SSM-style layer, and its per-sequence state is a fixed-size matrix that is updated recurrently as tokens arrive, in contrast to attention's KV cache, which grows linearly with sequence length. The two layer types therefore carry different kinds of state, and the cache layout, eviction policy, and sharing semantics differ between them.

\emph{Workload and metric.} A RAG-like workload in which every request shares a 32k-token system prefix, appends a 128-token request-specific suffix, and produces 128 output tokens. We report token-generation throughput at concurrency 20 on an NVIDIA L4 (24 GB), where memory pressure rules out keeping uncompressed per-request state copies.

\emph{Optimization opportunity.} Prefix sharing across requests is a standard technique, but a hybrid model needs two cache mechanisms in parallel: KV blocks for attention layers and a snapshot of the recurrent state at the prefix boundary for each DeltaNet layer. Knowing the workload at design time, the agent can specialize the system to this case and avoid the overhead of supporting prompt caching for generic workloads.

\emph{Why generic systems cannot exploit this.} vLLM and SGLang were architected around the attention KV cache. First-class hybrid-KV support is recent and limited~\cite{vllm-hybrid-kv,marconi}. Sharing the recurrent state across requests requires snapshotting at prefix boundaries, which incurs significant memory overhead to support generic cases, especially on hardware with limited memory capacity like L4. vLLM does expose a GDN-layer prefix-caching setting, \texttt{mamba\_cache\_mode}, which we enable in the baseline, but its two policies are both a poor fit for this workload's 32k-token shared prefix followed by a 128-token request-specific suffix: \texttt{"all"} snapshots the GDN state every 80 tokens (an interval sized from the relative sizes of the attention KV cache and the GDN state), producing a large number of snapshots that are never reused and overwhelm GPU memory capacity, while \texttt{"align"} snapshots only at the end of each prefill, which is also not reusable since the final 128 tokens differ across requests and the complete prefill is therefore never shared. This generic two-policy design cannot express the boundary this workload actually needs (a snapshot exactly at the 32k shared/unique split), which is what the bespoke system implements directly.

\paragraph{Scenario E: Streaming ASR with sliding-window encoder attention.}
\emph{Architecture.} Moonshine Streaming medium~\cite{kudlur2026moonshine} is an encoder-decoder ASR model designed for low-latency streaming. The encoder uses \emph{sliding-window} attention over audio frames, so previously encoded frames remain valid as new audio arrives; only the new tail needs to be encoded. The decoder is a small autoregressive Transformer that emits text tokens conditioned on encoder outputs. In contrast, Whisper~\cite{whisper}, the standard ASR baseline, encodes a full clip in a single pass and is not designed for incremental encoding.

\emph{Workload and metric.} 32 concurrent streaming clients, each sending a 2-second audio chunk every 2 seconds. We report time-to-first-token (TTFT) per chunk, which captures responsiveness for interactive transcription. We compare against a vLLM-plugin Moonshine baseline.

\emph{Optimization opportunity.} Sliding-window attention permits \emph{encoder-output caching}: each chunk encodes only the new tail and reuses the previous encoder outputs to feed the decoder. The system needs (i) a per-stream encoder cache aligned with the sliding window, (ii) eviction synchronized with the window's stride, and (iii) a scheduler that batches per-chunk encoder work alongside per-token decoder work across many concurrent streams.

\emph{Why generic systems cannot exploit this.} vLLM can support the Moonshine Streaming model via a plugin, but it cannot support encoder prompt caching without significant modifications to its code, which leads to redundant computation for streaming applications. In contrast, the bespoke system can optimize around Moonshine's specific sliding-window attention and expose the encoder layer to per-stream cache management.

\paragraph{Scenario F: Local constrained decoding for JSON generation.}
\emph{Architecture and target.} Llama-3.1-8B-Instruct~\cite{grattafiori2024llama} (8-bit MLX quantization) on a MacBook Pro (Apple M3 Pro, 36\,GB unified memory) running macOS 26.5 (build 25F5042g). We optimize single-stream end-to-end latency at $T{=}0$ on JSONSchemaBench~\citep{jsonschemabench}, a corpus of $\sim$9{,}558 real-world JSON schemas drawn from json-schema-corpus, GlaiveAI function-call schemas, and Kubernetes schemas, which is used to measure both the speed and the schema-feature coverage of constrained-decoding engines. The schemas are partitioned into 10 splits along two axes (domain and complexity): function-calling (GlaiveAI-2K, 1{,}707), operational/resource-access APIs (Snowplow 403, Washington Post 125), Kubernetes API configurations (1{,}064), a curated JSONSchemaStore set (492), and five GitHub-sourced \emph{Misc} tiers graded by constraint complexity (Trivial 444, Easy 1{,}943, Medium 1{,}976, Hard 1{,}240, Ultra 164). The distribution is skewed toward GitHub Easy/Medium and function-call schemas, with progressively rarer Hard/Ultra tails that stress less common JSON Schema features. The workload is held fixed across seven \sys{} iterations and we report p50 latency.

\emph{Optimization opportunity.} Three techniques compose. First, JSON-schema-constrained decoding with XGrammar~\cite{li2026xgrammar2efficientdynamicstructured} masks tokens that would violate the schema and applies \emph{jump-forward} to skip over deterministic tokens implied by the schema. Second, speculative decoding uses Llama-3.2-1B-Instruct (4-bit, MLX) as the draft against the 8B-8bit target with $K{=}4$ draft tokens per step; the smaller draft is preferred because its lower per-step cost outweighs its lower acceptance rate. Third, raising mlx\_lm's prefill chunk size from the default 512 to 2048 lets a $\sim$1300-token prompt prefill in a single chunk.

\emph{Why generic systems cannot exploit this.} vLLM and SGLang implement constrained decoding well, but only on CUDA backends. mlx\_lm runs on Apple Silicon and supports draft-model speculative decoding, but it has no support for grammar-constrained decoding, let alone a version that composes with speculative decoding. Beyond the missing backend, the wins here demand integration deeper than a generic structured-output API allows. The schema must be enforced inside the decoder via a per-token XGrammar bitmask with grammar-aware termination, and that bitmask must coexist with speculative decoding's rollback semantics: draft tokens may be partially accepted, so a constrained decoder that assumes monotonic token consumption silently drifts on rejection (\sys{} hit two such MLX correctness bugs during the study). The performance levers are similarly non-generic: \texttt{prefill\_step\_size{=}2048} is an MLX-specific fix for the interaction between $\sim$1300-token prompts and cache chunking; XGrammar's \texttt{any\_whitespace=False} and compact separators change token counts and thus latency; and several plausible generic optimizations (larger draft, different $K$, KV quantization, forced-token jump-forward) regressed in this configuration. Residual failures (unbounded \texttt{patternProperties}, nested arrays, approximate \texttt{oneOf}/\texttt{anyOf} unions) need schema-aware decoding behavior that off-the-shelf APIs do not expose. These failures reflect limitations of upstream XGrammar's JSON Schema support, not regressions introduced by \sys{}.

\paragraph{Scenario G: Local image generation with a unified vision-language model.}
\emph{Architecture.} Show-o2~\cite{showo2} is a unified vision-language model whose forward pass interleaves autoregressive (AR) text-token generation (a Qwen2.5-1.5B body) with diffusion-style image-token refinement (a 10-block diffusion head with SigLIP-based image conditioning). Each generation step is partly an AR decode (text body, with prefix-KV cache) and partly a denoising step (head, with classifier-free guidance over conditional and unconditional branches). The control flow does not match either a pure decoder-only LLM or a pure diffusion image model.

\emph{Workload and targets.} We evaluate text-to-image generation at $432{\times}432$ resolution with 20 sampler steps in two deployments. (i) MacBook Pro (Apple M3 Pro, 36\,GB unified memory) running macOS 26.5 (build 25F5042g), where we measure single-stream warm-min latency against the Show-o2 PyTorch-MPS reference implementation. (ii) NVIDIA H100, where we measure single-request latency at a fixed prompt and seed against the baseline's PyTorch implementation wrapped as a server. The benchmark client and server share the same container over the loopback interface and exchange raw PPM frames, so the measured target is serving latency rather than network or encoding overhead.

\emph{Accuracy gate.} Bitwise reproduction of the baseline is too restrictive when quantization, kernel substitution, or step-skipping is on the table. We provide \sys{} with a custom checker that compares each generated image against the baseline at a fixed prompt, seed, step count, guidance scale, and device profile; the checker accepts an image if it matches the baseline's $432{\times}432$ dimensions and meets a quality bar of MAE\,$\leq 2$, PSNR\,$\geq 35$\,dB, and local luminance SSIM\,$\geq 0.98$. The same checker gates both the H100 and MacBook variants.

\emph{Optimization opportunity.} \sys{} ports the body and head to the target backend (MLX on MacBook), elides a redundant SigLIP encode of noisy latents on every step, adds prefix-KV caches on body and head, trims prefill to the active image span, and applies a CFG \emph{stride} that skips the unconditional branch on $K{-}1$ of every $K$ denoising steps and reuses the cached \texttt{v\_uncond}. Quantization is restricted to the bandwidth-bound head; weight quantization on the compute-bound body regresses latency. On H100, aggressive trimming and naive batching regress quality. FlashAttention-2, GQA, \texttt{torch.compile}, and fp16 alter outputs or produce NaNs. Qwen prefix reuse yields no gain because the text prefix is tiny next to the 730-token image span. The $14.5$\,s MacBook floor replaces each per-step component with its fp16 kernel-perfect time. See \S\ref{sec:eval-show-o2} (H100) and \S\ref{sec:eval-show-o2-mac} (MacBook) for iteration-level breakdowns.

\emph{Why generic systems cannot exploit this.} There is no generic serving stack for Show-o2: vLLM does not implement diffusion paths, vLLM-Omni does not include this model, and the reference is a research-grade PyTorch implementation. The AR/diffusion interleaving and the body/head/sampler co-design needed for the wins above do not generalize across models, so adding Show-o2 to a generic stack would be model-specific work that competes for engineering attention with every other model the stack supports. A bespoke system can wire the loop around exactly this control flow.

\paragraph{Baseline configurations.} For reproducibility, we list the exact non-default flags used for each scenario's serving-system baseline.
\begin{itemize}[leftmargin=1.5em]
    \item \textbf{Scenario A:} vLLM with \texttt{--gpu-memory-utilization 0.95 --max-model-len 8192 --max-num-seqs 256 --max-num-batched-tokens 16384 --performance-mode throughput --optimization-level 2 --async-scheduling}.
    \item \textbf{Scenario B:} SGLang v0.5.18 with environment
    \path{SGLANG_USE_AITER=1}, \path{SGLANG_USE_AITER_UNIFIED_ATTN=1},
    \path{SGLANG_MXFP4_MOE_HIP=1}, \path{SGLANG_SKINNY_GEMM=1},
    \path{SGLANG_AITER_GEMM_PAD_M=64}, \path{SGLANG_MAMBA_SSM_DTYPE=bfloat16},
    \path{PYTORCH_TUNABLEOP_ENABLED=1}, and \path{PYTORCH_TUNABLEOP_TUNING=0},
    plus server flags \texttt{--tp 4 --attention-backend aiter --page-size 16 --mem-fraction-static 0.72 --max-total-tokens 787936 --speculative-algorithm NEXTN --speculative-eagle-topk 1 --speculative-num-steps 3 --speculative-num-draft-tokens 4}. Unlike the other scenarios, this baseline also required source-level changes to SGLang itself, not just flags. We describe those changes next.
    \item \textbf{Scenario C:} vLLM vanilla, plus \texttt{--gpu-memory-utilization 0.95 --max-model-len 4096}. The speculative-decoding baseline additionally sets \texttt{\{"method": "eagle3", "model": "AngelSlim/Qwen3-32B\_eagle3", "num\_speculative\_tokens": 7\}}.
    \item \textbf{Scenario D:} vLLM vanilla, plus \texttt{--dtype bfloat16 --max-model-len 36864 --gpu-memory-utilization 0.92 --enable-prefix-caching}.
    \item \textbf{Scenario E:} vLLM vanilla, with a port of Moonshine into vLLM as an out-of-tree model plugin.
    \item \textbf{Scenario F:} \texttt{mlx\_lm} with \texttt{--temp 0 --max-tokens 512 --decode-concurrency 1 --prompt-concurrency 1 --prompt-cache-size 0 --prefill-step-size 512}.
    \item \textbf{Scenario G:} the vanilla PyTorch reference implementation.
\end{itemize}

\paragraph{SGLang source modifications for Scenario B.} Scenario B's baseline is not stock SGLang plus flags, but SGLang with \sys{}-authored source changes. We summarize those changes at a high level for reproducibility. \sys{} replaced SGLang's gfx942 Triton MoE fallback with a custom fused HIP MXFP4 MoE kernel, added a skinny BF16 GEMM path for small-batch decode, and iteratively refined the MoE kernel with register byte permutation, wider packed-weight loads, and a lower stage-1 threshold. It also changed the serving path to reuse cached-prefix tokenization for the known turn-opening suffix instead of retokenizing it, and to use constant-time expert-weight mapping with threaded, sharded weight loading to reduce startup time (this last change does not affect steady-state throughput). These source changes sit on top of SGLang's existing NEXTN/MTP speculative-decoding support, which we enable through the flags above rather than modify.

\paragraph{Per-role agentic-loop breakdown.}
Table~\ref{tab:loop-breakdown} reports per-role LLM-call counts and active time across evaluation scenarios. Across all scenarios except B, the Implementer dominates active time in every run ($47$--$60\%$), reflecting that producing and revising candidate code is the most compute-intensive step. The Accuracy Judge is the next-largest contributor ($20$--$30\%$) and is especially heavy on Scenario~D, where the dual KV/recurrent-state cache machinery makes correctness review more involved. The Performance Evaluator runs less often because performance work is gated on a passing accuracy round, and the Orchestrator is consistently a small share ($3$--$7\%$) at low average agent execution time ($34$--$46$\,s) since it only selects the next issue and updates long-term memory.

Scenario B departs from this pattern. This is a hard model and hardware combination to optimize, so the Orchestrator reasons through multiple candidate optimization directions each round. That raises its average agent execution time ($574$\,s, $12$--$17\times$ the other scenarios) and its share of active time ($8.7\%$, above the range of the other scenarios). To save wall-clock time, the Orchestrator also dispatches multiple inner-loop rounds in parallel (\S\ref{sec:design:agent}), so independent optimizations proceed at their own pace and trigger correctness and performance measurements independently. These durations are therefore cumulative per-role active time that overlaps across parallel rounds. The $392.8$\,h of total active time corresponds to a $105$\,h end-to-end run, an average overlap of about $3.7\times$. The Performance Evaluator is the largest share of active time ($49.7\%$) and its average agent execution time ($2164$\,s) is well above the other scenarios ($175$--$373$\,s). Booting a distributed serving engine for a model this large takes longer, and we run on a shared GPU cluster with unpredictable contention, so evaluation jobs can also wait in the queue. Judge and Implementer average agent execution times ($1080$\,s and $1205$\,s) are also above the other scenarios ($139$--$411$\,s and $393$--$648$\,s).

\paragraph{Cost.} Table~\ref{tab:token-cost} reports token usage and cost per run. We ran all experiments on Codex or Claude subscriptions, so the token costs in the table are not what we were billed. They are computed from token consumption at API list prices, as an estimate of what each run would cost under per-token billing. Because agents repeatedly re-read long, mostly unchanged contexts, 94--99\% of input tokens are served from the prompt cache. In Scenarios~A--G, these cached reads account for most of the token cost: \$3{,}150 of the \$4{,}919 total (64\%), and more than half of the cost in every run except the SGLang tuning for Scenario~B (42\%). Even with GPU time included, the runs are not prohibitively expensive. Scenarios~A and C--G each cost a few hundred dollars (\$90--\$744). The two Qwen3.6 MI250 runs of Appendix~\ref{app:extend-sglang} cost \$1{,}348 (bespoke engine) and \$980 (SGLang). Scenario~B is the outlier, at about \$3{,}900 including its SGLang tuning run, because it is the longest run, consumes the most tokens, and occupies MI300A hardware for over 100 hours. The SGLang-seeded Scenario~B campaign (Appendix~\ref{app:extend-sglang-b}) cost about \$2{,}725.

\begin{table}[h]
\centering
\footnotesize
\caption{Per-role LLM-call breakdown across evaluation scenarios. ``Calls'' counts agent invocations, ``Duration'' is cumulative active time, ``Share'' is the fraction of the scenario's total active time, and ``Avg/call'' is the mean agent execution time per invocation, from receiving a task to delivering its result. The parenthesis after each scenario is its end-to-end wall-clock time. Roles can run in parallel, so per-role durations overlap and their sum can exceed it. Roles correspond to the inner-loop Implementer, Accuracy Judge, and Performance Evaluator (\S\ref{sec:design}) plus the outer-loop Orchestrator.}
\label{tab:loop-breakdown}
\setlength{\tabcolsep}{4pt}
\begin{tabular}{@{}llrrrr@{}}
\toprule
Scenario & Role & Calls & Duration (h) & Share & Avg/call (s) \\
\midrule
\multirow{4}{*}{A: Llama-3.1-8B standard (25.0\,h)}
  & Orchestrator & 120 & 1.40 & 5.6\%  & 42 \\
  & Implementer  & 90  & 13.35 & 53.4\% & 534 \\
  & Judge        & 90  & 6.08 & 24.3\% & 243 \\
  & Perf.\ Evaluator & 60  & 4.18 & 16.7\% & 251 \\
\midrule
\multirow{4}{*}{B: AMD distributed MoE (105\,h)}
  & Orchestrator & 215 & 34.3 & 8.7\%  & 574 \\
  & Implementer  & 396 & 132.5 & 33.7\% & 1205 \\
  & Judge        & 102 & 30.6 & 7.8\%  & 1080 \\
  & Perf.\ Evaluator & 325 & 195.4 & 49.7\% & 2164 \\
\midrule
\multirow{4}{*}{C: Qwen3-32B code edit (8.66\,h)}
  & Orchestrator & 47 & 0.61 & 7.0\%  & 46 \\
  & Implementer  & 36 & 4.55 & 52.5\% & 455 \\
  & Judge        & 36 & 1.98 & 22.8\% & 198 \\
  & Perf.\ Evaluator & 23 & 1.53 & 17.6\% & 239 \\
\midrule
\multirow{4}{*}{D: Olmo-Hybrid prefix caching (12.54\,h)}
  & Orchestrator & 47 & 0.44 & 3.5\%  & 34 \\
  & Implementer  & 33 & 5.94 & 47.4\% & 648 \\
  & Judge        & 33 & 3.77 & 30.1\% & 411 \\
  & Perf.\ Evaluator & 23 & 2.39 & 19.1\% & 373 \\
\midrule
\multirow{4}{*}{E: Moonshine streaming (8.71\,h)}
  & Orchestrator & 49 & 0.55 & 6.3\%  & 41 \\
  & Implementer  & 46 & 5.25 & 60.3\% & 411 \\
  & Judge        & 45 & 1.74 & 20.0\% & 139 \\
  & Perf.\ Evaluator & 24 & 1.17 & 13.4\% & 175 \\
\midrule
\multirow{4}{*}{F: JSON constrained decoding (3.0\,h)}
  & Orchestrator & 11 & 0.13 & 4.3\%  & 43 \\
  & Implementer  & 13 & 1.42 & 47.3\% & 393 \\
  & Judge        & 13 & 0.85 & 28.3\% & 235 \\
  & Perf.\ Evaluator & 7  & 0.60 & 20.0\% & 309 \\
\midrule
\multirow{4}{*}{G: Show-o2 H100+MBP (14.0\,h)}
  & Orchestrator & 70 & 0.85 & 6.1\%  & 44 \\
  & Implementer  & 53 & 6.85 & 48.9\% & 465 \\
  & Judge        & 51 & 3.05 & 21.8\% & 215 \\
  & Perf.\ Evaluator & 34 & 3.25 & 23.2\% & 344 \\
\bottomrule
\end{tabular}
\end{table}

\begin{table}[h]
\centering
\footnotesize
\caption{Token usage and cost per run, summed over all roles. ``Cache hit'' is the fraction of input tokens served from the prompt cache. Token cost prices the tokens at API list rates. GPU cost uses Modal's on-demand rates for H100 (\$3.95/h) and L4 (\$0.80/h), and estimates of \$8/h for MI300A and \$7 per node-hour for MI250 (one node is eight GCDs). Node-hours are summed over concurrent jobs, so they can exceed wall-clock time. MacBook runs are unbilled. Dollars are rounded.}
\label{tab:token-cost}
\setlength{\tabcolsep}{4pt}
\begin{tabular}{@{}lrrrrlrr@{}}
\toprule
 & \multicolumn{4}{c}{Tokens} & \multicolumn{3}{c}{GPU} \\
\cmidrule(lr){2-5}\cmidrule(l){6-8}
Scenario & Input (M) & Cache hit & Output (M) & Cost (\$) & Device & Hours & Cost (\$) \\
\midrule
A: Llama-3.1-8B & 940     & 96.0\% & 2.2  & 703   & H100    & 10.4  & 41 \\
B: AMD MoE & 3{,}280 & 98.3\% & 21.8 & 2{,}425 & MI300A & 108 & 864 \\
\quad SGLang tuning     & 950     & 94.0\% & 0.8  & 433   & MI300A  & 19    & 152 \\
C: Qwen3-32B & 378.6   & 97.4\% & 0.71 & 255   & H100    & 8.7   & 34 \\
D: Olmo-Hybrid          & 320.2   & 96.8\% & 0.54 & 222   & L4      & 12.5  & 10 \\
E: Moonshine & 430.1   & 97.4\% & 0.73 & 287   & L4      & 8.7   & 7 \\
F: JSON decoding        & 122.4   & 98.1\% & 0.60 & 90    & MacBook & --    & -- \\
G: Show-o2 (H100)       & 477.4   & 97.8\% & 0.67 & 305   & H100    & 3.5   & 14 \\
G: Show-o2 (Mac) & 264.9   & 95.9\% & 0.61 & 200   & MacBook & --    & -- \\
\midrule
Qwen3.6 MI250 (bespoke) & 1{,}702 & 98.0\% & 9.6 & 879 & MI250 & 67 & 469 \\
Qwen3.6 MI250 (SGLang)  & 746     & 99.0\% & 3.5 & 441 & MI250 & 77 & 539 \\
B: AMD MoE (SGLang seed) & 2{,}560 & 98.0\% & 5.4 & 1{,}733 & MI300A & 124 & 992 \\
\bottomrule
\end{tabular}

\end{table}

\section{Ablation Details}
\label{app:extend-vllm}

This appendix gives the full comparison tables and per-round detail for the two ablations summarized in \S\ref{sec:eval-ablation}.

\paragraph{Starting-point ablation.} \sys is seeded from vLLM v0.26.0 (Scenarios D, E) or vLLM-Omni v0.24.1 (Scenario G), or SGLang (the MI250 task below), instead of an empty repository, and compared against the from-scratch runs of \S\ref{sec:eval-results}.

\begin{table}[h]
\centering
\footnotesize
\caption{Ablating the starting point: seeding \sys with an existing codebase instead of building from scratch. Outcomes are mixed: convergence accelerates when the seed's abstractions already fit the target (D), and the agent works around the seed's abstractions otherwise (E, G).}
\label{tab:extend-existing}
\setlength{\tabcolsep}{4pt}
\begin{tabular}{@{}p{0.30\linewidth}p{0.20\linewidth}p{0.36\linewidth}@{}}
\toprule
Scenario & Seed & Outcome \\
\midrule
D: Olmo-Hybrid prompt caching & vLLM v0.26.0 & Matches from-scratch parity by round~4 ($\approx\!2\times$ faster) \\
E: Moonshine streaming ASR & vLLM v0.26.0 & Agent implements outside vLLM's abstractions \\
G: Show-o2 (H100) & vLLM-Omni v0.24.1 & Agent implements outside vLLM-Omni's abstractions \\
Qwen3.6-35B-A3B (MI250) & SGLang & SGLang agents lead early, but bespoke engine overtakes eventually ($1{,}375$ vs.\ $815$ tok/s) \\
Qwen3.5-397B-A17B (MI300A), Scenario B & SGLang & SGLang agents lead briefly (${\sim}50$--$72$\,h), then plateau; bespoke engine ends $1.37\times$ higher at equal budget ($2{,}242$ vs.\ $1{,}637$ tok/s) \\
\bottomrule
\end{tabular}
\end{table}

Per-round detail follows for each seeded scenario.

\paragraph{Scenario D: Olmo-Hybrid prompt caching.} The agent patches vLLM's OLMo-Hybrid fused conv1d weight loader (a shard-id-recovery bug after \texttt{WeightsMapper.apply()}), configures the V1 engine (\texttt{AsyncLLM}, chunked prefill, \texttt{FULL\_AND\_PIECEWISE} CUDA graphs), and replaces a per-request global lock with vLLM's native continuous batching. Table~\ref{tab:extend-olmo-rounds} traces throughput on the same 20-concurrent-request, 32k-shared-prefix workload as \S\ref{sec:eval-olmo-hybrid} across 8 rounds. A controlled A/B at round~5 isolates the effect of prefix caching: enabling it raises throughput from 6.04 to 74.98 tok/s ($12.4\times$) at matched request count, confirming that the seeded run's gains come from prefix reuse rather than incidental engine tuning.

\begin{table}[h]
\centering
\footnotesize
\caption{Scenario D (Olmo-Hybrid, seeded from vLLM v0.26.0): per-round throughput on the 20-request, 32k-shared-prefix workload.}
\label{tab:extend-olmo-rounds}
\begin{tabular}{@{}llr@{}}
\toprule
Round & What changed & Throughput (tok/s) \\
\midrule
R2 (cold) & Serving-contract baseline & 8.66 \\
R2 (hot) & Same, prefix-warm & 10.69 \\
R3 & Profiler/tokenizer fix & 10.33 \\
R4 & Endpoint fix, 20/20 requests stable & 76.01 \\
R5 (cache enabled) & Prefix-cache A/B, enabled & 74.98 \\
R5 (cache bypassed) & Prefix-cache A/B, bypassed & 6.04 \\
Hot final & Best hot measurement & 85.98 \\
R7 & Profiler contract repair & 75.58 \\
R8 & Worker-side CUDA profiling & 75.86 \\
\bottomrule
\end{tabular}
\end{table}

\paragraph{Scenario E: Moonshine streaming ASR.} vLLM v0.26.0 is present in the environment but unused for inference (vLLM has no Moonshine model), so the agent implements a custom encoder-decoder path outside it, matching the outcome in Table~\ref{tab:extend-existing}. Table~\ref{tab:extend-moonshine-rounds} traces TTFT p50 over 2-second streaming audio chunks. The jump from round~5 to rounds~6--7 (5902\,ms $\to$ 430\,ms) comes from moving to concurrent streaming with bounded per-session admission and a queue/drain microbatcher, not from model-level changes. Per-chunk cost is dominated by the decoder ($\sim$464\,ms) rather than the encoder ($\sim$19\,ms).

\begin{table}[h]
\centering
\footnotesize
\caption{Scenario E (Moonshine, seeded from vLLM v0.26.0): per-round streaming TTFT p50.}
\label{tab:extend-moonshine-rounds}
\begin{tabular}{@{}llrr@{}}
\toprule
Round & What changed & TTFT p50 (ms) & Concurrency \\
\midrule
R2 & Baseline serving contract & 7076.6 & 1 \\
R4 & Streaming encoder partial fast path & 6079.5 & 1 \\
R5 & Decoder KV/cross-attention cache & 5902.0 & 1 \\
R6--7 & Bounded admission, disconnect-safe streaming & 430.2 & 4 \\
R8 & Queue/drain microbatcher & 367.0 & 4 \\
R9 & Encoder CUDA-graph buckets (warm) & 293.8 & 4 \\
R11 & Recovery to R9 path & 295.4 & 4 \\
R13 & Concurrent-function cohort wait & 306.2 & 4 \\
Final & Best stable warm & 268.4 & 4 \\
\bottomrule
\end{tabular}
\end{table}

\paragraph{Scenario G: Show-o2 (H100).} The agent builds a serving layer around vLLM-Omni (FastAPI endpoints, VAE integration) but implements Show-o2's active-span forward pass --- projecting only the 4,380 active image-latent tokens instead of the full 6,144-token sequence --- outside vLLM-Omni's model-serving abstractions, again matching Table~\ref{tab:extend-existing}. Table~\ref{tab:extend-showo2-rounds} traces single-request p50 latency at 4 diffusion steps over 13 rounds. The largest single gain (509.6\,ms $\to$ 428.6\,ms) comes from the active-span path, and response-serialization tuning contributes a further $\sim$104\,ms.

\begin{table}[h]
\centering
\footnotesize
\caption{Scenario G (Show-o2, seeded from vLLM-Omni v0.24.1): per-round T2I p50 latency, 4 diffusion steps.}
\label{tab:extend-showo2-rounds}
\begin{tabular}{@{}llr@{}}
\toprule
Round & What changed & p50 latency (ms) \\
\midrule
Baseline & Server contract established & 509.6 \\
R6 & Active-span fast T2I path (4380/6144 tokens) & 428.6 \\
R11 & ASGI benchmark forwarder & 396.1 \\
R13 & Pre-rendered JSON, lower-copy base64 & 324.8 \\
\bottomrule
\end{tabular}
\end{table}

\begin{figure}[h]
\centering
\includegraphics[width=0.95\linewidth]{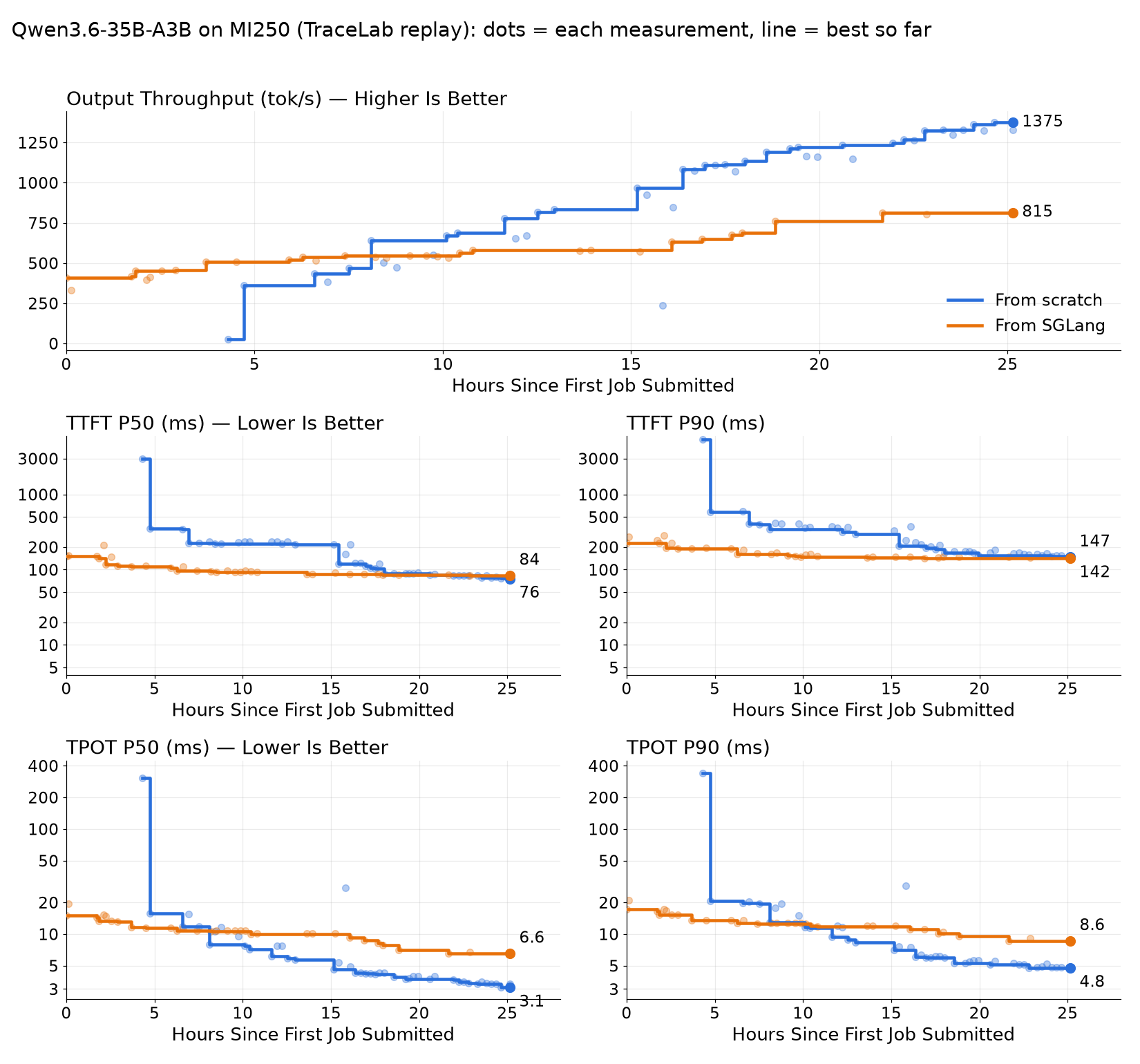}
\caption{Each dot is one measurement and each line is the best value so far. Time zero is the first measurement of either run. Microbenchmark runs not plotted.}
\label{fig:extend-sglang}
\end{figure}

\paragraph{Qwen3.6-35B-A3B on MI250: agents optimizing SGLang vs.\ a bespoke engine from scratch.}\label{app:extend-sglang}
We ran two \sys{} campaigns comparing starting from SGLang vs. from scratch. The runs used Claude Code Opus 5.5 as agents, with the same time budget, on Qwen3.6-35B-A3B (BF16), on one node with eight GPUs (four MI250 cards connected by XGMI, each with two GPUs linked in-package), replaying TraceLab~\cite{tracelab} coding agent traces. SGLang first had to be ported to run on gfx90a. About 25 hours after the first measurement, the agents optimizing the bespoke engine (14k lines of source) reached 1{,}375 tok/s against 815 for the agents optimizing SGLang (4.1k lines of source added) (Table~\ref{tab:extend-sglang}, Figure~\ref{fig:extend-sglang}). The bespoke run also used more tokens: 1.7B input and 9.6M output tokens (\$879 at API prices) against 0.75B and 3.5M (\$441) for the SGLang run, but fewer GPU node-hours (67 against 77; Table~\ref{tab:token-cost}).

The SGLang agents made steady progress with changes that stayed inside one operation. They enabled prefill graph capture, which SGLang supports but had off for HIP (a five-line change, TTFT p50 from 143 to 113\,ms). They wrote a better small-batch GEMM kernel and a fused kernel for the recurrent layer's decode step, tuned the MoE configurations, and made several other kernel changes. Together, these raised throughput from 408\,tok/s, right after the layout change, to 815. However, the performance plateaued after that. Agents wrote microbenchmarks to prototype bigger changes that cut across multiple operations, but did not lead to end-to-end performance gains.

The bespoke engine, on the other hand, made three optimizations that each span several components. SGLang either lacks them or its agents tried them but did not pay off.
\begin{itemize}[leftmargin=*,itemsep=2pt,topsep=2pt]
\item \textbf{Whole-layer decode fusion.} A decode layer originally ran dozens of tiny kernels, each writing its result to HBM before the next read it. The bespoke agents merged the input norm, the four input projections with the convolution, the recurrent update, the output projection, the router and expert layers, and the residual adds into about ten kernels per layer. This cut the two-request step from 27 to 11.5\,ms and raised throughput in a replay-like test from 236 to 451\,tok/s. In SGLang, a layer is assembled from separate components (the all-reduce, the linear, MoE and recurrent layers, and graph capture), so a fused layer has to take over all of them. The SGLang agents fused kernels in pairs, each raising throughput by 6--7\%. Late in the run they landed the first stage of a persistent layer kernel, covering the router and expert layers, which raised throughput by 13\% (673 to 761\,tok/s) and cut the kernels per step from 868 to 458. Tuning that kernel brought it to 815. The bespoke agents did the whole layer at once because they owned its forward code end-to-end.
\item \textbf{Speculative decoding with double-buffered recurrent state.} The model's MTP head drafts the next token, and the target model verifies it in the same step. The recurrent state is overwritten by every token, so the bespoke engine keeps two copies per request and switches to the spare only if the draft is accepted. This raised throughput by about 40\% (236 to 328\,tok/s in a replay-like test), because a verify step cost only about $1.25\times$ a decode step. In SGLang, the same change touches the state memory pool, the recurrent kernels of each backend, the accept logic and graph capture of the draft steps. The SGLang agents attempted it but failed twice, first because output was non-repeatable and the fix cancelled the gain, then because a verify step cost $2.6\times$ a decode step.
\item \textbf{A fused, fully graph-captured prefill pass.} The bespoke agents merged the norms, rotary embedding, gating and recurrent-layer setup, then captured the whole pass as one replayable graph, which cut TTFT p50 from 214 to 122\,ms. SGLang's main implementation replays only part of the pass. Its attention and recurrent layers still launch their kernels one by one from the CPU, because their launch sizes change with each batch.
\end{itemize}

The bespoke agents also changed serving policy, which is easy when one agent owns the request loop. They prefilled each round's uncached prompt in one pass instead of fixed chunks (TTFT p50 from 346 to 224\,ms, and throughput up 7\%). They made the speculation depth adaptive, up to seven tokens ahead, which raised throughput by 16\%. They pinned each replica's host threads to the CPU socket local to its GPUs (TTFT p90 down 7\%). The SGLang agents tuned routing and cache settings, but their plans include no comparable change to batching or speculation policy, which in SGLang belong to a scheduler shared by every workload. In short, SGLang's existing abstractions made it much more difficult for agents to find, design, and implement the optimizations that work well for this (model, hardware, workload) combination.

\begin{table}[h]
\centering
\footnotesize
\caption{Best result about 25 hours after the first measurement, Qwen3.6-35B-A3B on MI250.}
\label{tab:extend-sglang}
\begin{tabular}{@{}lrr@{}}
\toprule
 & Bespoke engine & SGLang \\
\midrule
Output tok/s & 1{,}375 & 815 \\
TTFT p50 / p90 (ms) & 76 / 147 & 84 / 142 \\
TPOT p50 / p90 (ms) & 3.1 / 4.8 & 6.6 / 8.6 \\
\bottomrule
\end{tabular}

\end{table}

\begin{figure}[h]
\centering
\includegraphics[width=0.95\linewidth]{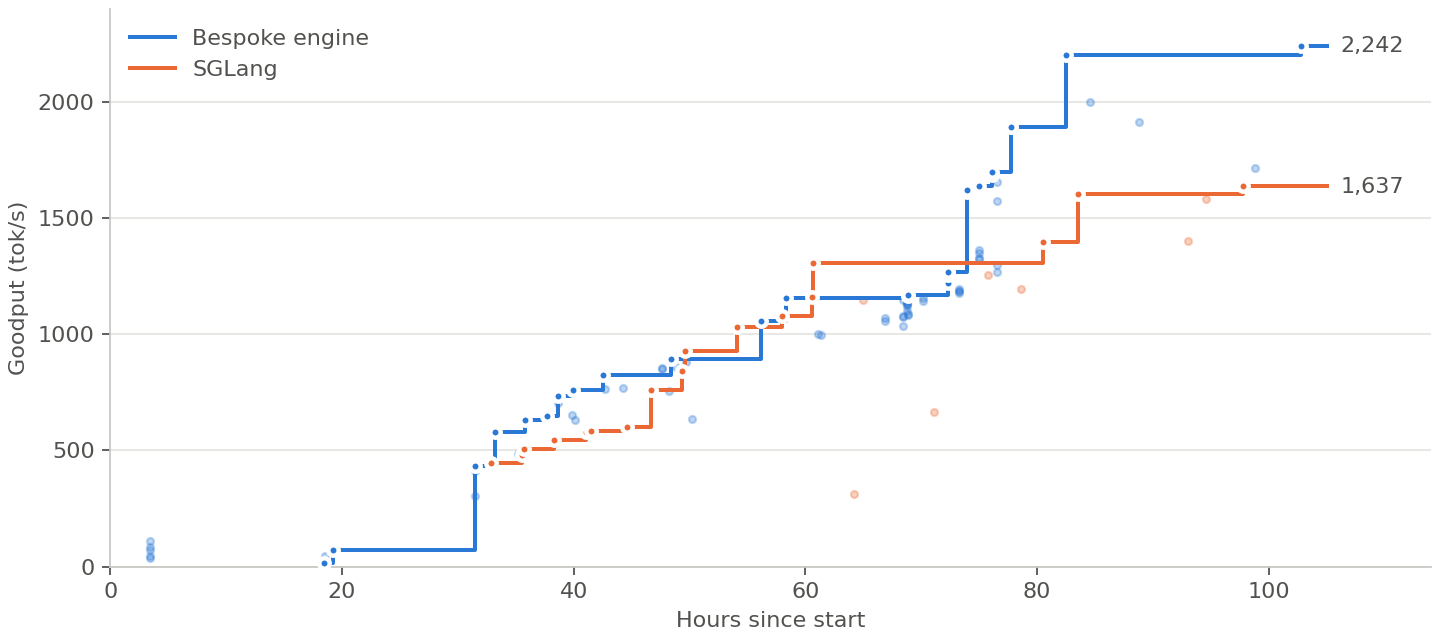}
\caption{Scenario B (Qwen3.5-397B-A17B-MXFP4 on 4$\times$ MI300A): best goodput so far for the bespoke engine and for the SGLang-seeded campaign, against hours since start. Dots are individual measurements. The bespoke curve is the original run of the main paper, replotted against hours since start. Both campaigns have a 105-hour budget, and SGLang's porting time is included.}
\label{fig:extend-sglang-b}
\end{figure}

\paragraph{Scenario B: agents optimizing SGLang vs.\ a bespoke engine at an equal budget.}\label{app:extend-sglang-b}
The $2.33\times$ gain of \S\ref{sec:eval-amd-dist} is measured against an SGLang that \sys{} ported and tuned for a smaller budget. To compare the two starting points fairly, we ran a new \sys{} campaign that starts from SGLang on Scenario~B, with the same agents, skills library, accuracy checker, workload, and 105-hour budget as the bespoke run of the main paper. SGLang does not run this model on this hardware out of the box (\S\ref{app:eval-scenarios}), so the campaign first had to port it, and the porting time counts toward the 105 hours. Figure~\ref{fig:extend-sglang-b} plots both campaigns against hours since start. The bespoke curve is the original run from the paper.

At 96 concurrent sessions, the highest concurrency before overload, the bespoke engine reaches $2{,}242$\,tok/s. The best measurement from the SGLang-seeded campaign is $1{,}637$\,tok/s, so the bespoke engine is $1.37\times$ higher at an equal time budget.

The two campaigns track each other closely for the first $\sim$60 hours, since SGLang also had to be ported and optimized for this hardware. SGLang was briefly ahead between $\sim$50\,h and $\sim$72\,h ($1{,}305$ vs.\ $\sim$$1{,}155$\,tok/s at $\sim$61\,h). The bespoke run then overtook it at $\sim$72\,h and took a series of larger steps (about $1{,}620$\,tok/s at $\sim$74\,h, $1{,}890$ at $\sim$78\,h, and $2{,}200$ at $\sim$83\,h). The SGLang-seeded campaign plateaued after $\sim$83\,h at about $1{,}600$--$1{,}640$\,tok/s.

Most of the gap comes from the following:
\begin{itemize}[leftmargin=*,itemsep=2pt,topsep=2pt]
\item \textbf{Speculative decoding (MTP).} The bespoke engine implements MTP speculative decoding, whereas the SGLang port does not, which is worth $+16$--$20\%$. The SGLang agents tried twice to land it and failed. The first attempt did not pass the task's accuracy checks, and the second implementation attempt gave no speedup.
\item \textbf{Sequence-parallel all-reduce.} The bespoke engine implements it to reduce the time spent in all-reduce.
\item \textbf{More aggressive kernel fusion.} The bespoke agents fused prefill kernels (in-place DeltaNet state, fused DeltaNet norm, fused prefill attention), RoPE with KV writes, routing, elementwise and norm operations, and the all-reduce with the residual add and RMSNorm. The SGLang agents attempted far less fusion.
\end{itemize}
As in the MI250 experiment, the SGLang agents made progress inside individual operations, while the bespoke engine landed optimizations crossing multiple components. The takeaway matches the MI250 experiment: agents make cross-cutting optimizations more easily in a bespoke engine, where they own the whole forward path and request loop, than in a generic framework whose abstractions must serve a wide range of hardware, models, and workloads.

The SGLang-seeded campaign cost about \$2{,}725 at API list prices, against \$3{,}289 for the bespoke run (Table~\ref{tab:token-cost}), so the bespoke engine delivered $1.37\times$ the goodput for about $1.2\times$ the cost.

\paragraph{Agentic pipeline ablation.} Table~\ref{tab:ablation} gives the full comparison across all four pipeline configurations on Scenario A (Llama-3.1-8B, H100, Codex GPT-5.5 Medium); \S\ref{sec:eval-ablation} discusses each ablation's failure mode.

\begin{table}[h]
\centering
\footnotesize
\caption{Ablating the agentic pipeline on Scenario A (Llama-3.1-8B, H100), Codex GPT-5.5 Medium. Each ablation removes one design choice from the full \sys pipeline; none converges to a competitive serving system.}
\label{tab:ablation}
\setlength{\tabcolsep}{4pt}
\begin{tabular}{@{}p{0.21\linewidth}p{0.13\linewidth}p{0.19\linewidth}p{0.39\linewidth}@{}}
\toprule
Configuration & Peak tok/s & Budget & Failure mode \\
\midrule
Single-agent & 617 & 35\,h, 1 context & Stuck in local optimization; skips accuracy/benchmark gates \\
No-skill & 1{,}100 & $\sim$35 rounds, 2 days & Exhausts incrementally discoverable optimizations \\
Outer-loop $\to$ OpenEvolve & 537 & 30 rounds, 1 day & Combines early wins, misses full optimization set \\
Full \sys & $\approx$10k (parity) & 60 rounds, 25.0\,h & --- \\
\bottomrule
\end{tabular}
\end{table}

\paragraph{Additional ablation: the first 15 rounds.}\label{app:ablation-15}
To separate the contributions of the Orchestrator's planning state, we reran Scenario A with Codex GPT-5.6-Sol on the same task, benchmark and accuracy check, and compared five configurations over their first 15 rounds. These results are not directly comparable to those in Table~\ref{tab:ablation} because the two experiments used different models. Besides the full pipeline, the no-skill run and OpenEvolve, two new ablations remove planning state from the Orchestrator. \emph{No issue tracker}: the Orchestrator neither reads nor writes the issue roadmap but keeps the round history. \emph{No long-term memory}: the round history, plateau summary and pending claims are removed as well, and only the best passed idea, the current frontier and the latest profile remain. For OpenEvolve, a round is one generation of two child candidates, so it evaluates about twice as many candidates per round. Table~\ref{tab:ablation-15} gives the best accepted throughput in 15 rounds.

\begin{table}[h]
\centering
\footnotesize
\caption{Best accepted throughput so far (tok/s) over the first 15 rounds, Scenario A (Llama-3.1-8B, H100). ``---'': no accepted candidate yet.}
\label{tab:ablation-15}
\setlength{\tabcolsep}{4pt}
\begin{tabular}{@{}cccccc@{}}
\toprule
Round & Full \sys & No skill & OpenEvolve & No issue tracker & No memory \\
\midrule
1 & 1{,}611 & --- & 2{,}750 & 178 & 3{,}467 \\
3 & 3{,}350 & 564 & 2{,}815 & 1{,}334 & 4{,}119 \\
5 & 5{,}553 & 1{,}156 & 2{,}869 & 1{,}519 & 4{,}125 \\
7 & 6{,}963 & 1{,}156 & 2{,}869 & 1{,}554 & 5{,}111 \\
10 & 6{,}968 & 1{,}156 & 2{,}996 & 1{,}554 & 5{,}111 \\
13 & 7{,}374 & 1{,}156 & 3{,}159 & 1{,}554 & 5{,}111 \\
15 & \textbf{7{,}374} & 1{,}156 & 3{,}180 & 4{,}395 & 5{,}698 \\
\bottomrule
\end{tabular}

\end{table}

The full pipeline climbs fastest and most steadily. It goes from 1{,}611 to 6{,}950~tok/s by round~6 and reaches 7{,}374 by round~13, and 13 of its 15 rounds produce an accepted, measured candidate. Without long-term memory, the run starts strongly (3{,}467~tok/s after round~1, 5{,}111 by round~7), but then goes seven rounds without improving and ends at 5{,}698. Without the issue tracker, the run stalls hardest. After round~6 its best stays at 1{,}554~tok/s for eight consecutive rounds. Two of those rounds measured above 4{,}300~tok/s but failed the Judge, and only round~15 yields an accepted, measured improvement, at 4{,}395. This is consistent with the issue tracker's role in separating implementation failures from unsuitable directions (\S\ref{sec:design:agent}). The no-skill run plateaus at 1{,}156~tok/s from round~5, close to the 1{,}100~tok/s of Table~\ref{tab:ablation}. OpenEvolve improves in small steps from its first candidate (2{,}739) to 3{,}180, and is ahead of only the no-skill run at round~15.

\section{Skills Library Construction}
\label{app:skills}

This appendix describes how the serving-systems skills library of \S\ref{sec:design:skills} is built, organized, and delivered to agents. The library changes along with \sys. The counts below describe it at the time of writing, and each run copies a snapshot of it into the agents' workspace at launch.

\paragraph{One skill, read on demand.}
Agent harnesses load the name and description of every installed skill into context at startup~\citep{agentskills}. Packaging each of the library's roughly 50 topics as its own skill would add about 5k tokens to every agent's context. The library is therefore a single skill whose description is about 400 characters of trigger terms, such as KV cache, speculative decoding, and MLX. The skill's body only routes. It tells the agent to open its platform's directory first, then to open the one or two references that match the current task, and it lists each portable reference in one line. In total the library holds 82 reference files (about 12.7k lines), which agents open with ordinary file reads. The skill also includes an overview of performance foundations, such as the roofline model and why prefill is compute-bound while decode is memory-bound. The overview maps symptoms, such as slow decode or a full KV cache, to the relevant entries.

\paragraph{Tiers.}
References are grouped by the abstraction layer they describe (Table~\ref{tab:skills-tiers}). Backend libraries and hardware are stored together, with one directory per compute backend, because which kernel libraries exist depends on the hardware.

\begin{table}[h]
\centering
\footnotesize
\caption[Tiers of the serving-systems skills library.]{Tiers of the serving-systems skills library. Platform files are counted across all five backends; a run sees only its own backend's directory.}
\label{tab:skills-tiers}
\setlength{\tabcolsep}{4pt}
\begin{tabular}{@{}p{0.13\linewidth}r p{0.34\linewidth}p{0.42\linewidth}@{}}
\toprule
Tier & Files & Content & Examples \\
\midrule
Models & 10 & Architecture families: weight layout, attention type, pitfalls specific to each modality & Dense and MoE text models, SSM hybrids, vision-language, speech, image and video generation, attention variants \\
Algorithms & 15 & Portable contracts: invariants, failure modes, compatibility table & Continuous batching, paged attention, radix and hybrid-model prefix caching, speculative decoding, structured output, MoE dispatch, parallelism \\
Frameworks & 2 & Programming idioms that work on every platform & PyTorch, Triton \\
Platforms & 41 & Per backend: default optimizations, hardware, profilers, kernel libraries & CUDA (FlashInfer, FlashAttention, CUDA graphs), ROCm (AITER), Apple Silicon (MLX), Trainium (NKI), CPU \\
Engines & 4 & Maps of where things live in reference engines' source & vLLM, SGLang, TensorRT-LLM \\
Tooling & 10 & Workflows separate from the serving design itself & Accuracy checker, serving benchmark, profiling, performance modeling, OpenAI-compatible API, I/O handling, LoRA \\
\bottomrule
\end{tabular}%

\end{table}

\paragraph{Sources.}
Entries are distilled from the source code of vLLM, SGLang, and TensorRT-LLM~\citep{pagedattention,sglang,tensorrt-llm}, from the documentation of kernel libraries such as FlashInfer and FlashAttention~\citep{flashinfer,dao2022flashattention}, and from the serving literature. The engine entries map where things live in each engine. Their ``where is X'' tables give the path of a component in each engine, such as the scheduler, an attention backend, or the speculative-decoding workers. An agent can then study how a production engine implements a mechanism without reading the whole codebase. Algorithm and model entries summarize the production implementations they draw on. For example, the asynchronous-scheduling entry describes SGLang's overlap scheduler and vLLM's asynchronous scheduler as they appear in the source. Each entry's cited paths were checked against copies of the engines' source when the entry was written. Entries were also checked against upstream documentation, which caught errors such as a misnamed FlashInfer MLA wrapper. The engine source code is kept outside the skill and is not copied into agent workspaces.

\paragraph{Entry structure.}
Each reference is a self-contained Markdown file of at most 500 lines. Its sections come in a fixed order: purpose, prerequisites, concept, workflow (code outlines and checklists), a compatibility table or source map, pitfalls, and further reading. Longer topics are split into linked sub-files. Algorithm entries are written as contracts. Each one states the problem, the invariants that any implementation must hold, and a table that maps observed symptoms to the broken invariant. For continuous batching, the invariants include positions and attention masks kept per request, admitting new requests only between steps, and admitting a request only when the KV cache has room for it. The symptom ``correct at batch size 1, wrong at larger batches'' points to a masking error. Each algorithm entry ends with a compatibility table that covers implementation, engine, kernel library, and hardware backend. The table includes explicit N/A rows, because a missing row would let a backend silently pick up another backend's guidance.

\paragraph{Portability across hardware.}
Guidance that is correct on one backend can be wrong on another. On GPUs, continuous batching removes padding from attention. On Trainium, whose compiler requires static shapes, padding inputs up to fixed sizes is the correct choice. Every topic is therefore classified by one test: would following it on another backend produce wrong work?
\begin{itemize}[leftmargin=*,nosep]
\item A portable topic keeps a single copy, and platform differences appear only as rows in its compatibility table.
\item When the problem is shared but the technique differs, the topic is split into a portable contract plus one implementation file per backend.
\item Material that exists on only one platform, such as a specific kernel library, profiler, or instruction set, lives only under that backend.
\end{itemize}
Each backend directory (CUDA, ROCm, Apple Silicon, Trainium, and CPU) must contain three files. The first lists the optimizations that should be on by default on that hardware. For example, CUDA graphs are required on NVIDIA GPUs and do not exist on Apple Silicon. The second covers the hardware's specifications, and the third covers its profilers and how to measure reliably. Notes on that backend's kernel libraries sit beside them.

\paragraph{Delivery to agents.}
When a run starts, \sys copies the skill into the agents' workspace and keeps only the target backend's platform directory, so an agent never reads another platform's guidance. Automated checks make sure this removal leaves the library usable. They reject a backend directory that lacks one of its three required files or is not named after a supported backend. They also reject links from portable files into a platform directory, because those links would break once other platforms are removed. A scoping rule loads the library only for LLM-serving targets.

\paragraph{Extension.}
Adding a model family means adding one file to the models tier. Adding a backend takes three steps: (1)~create its directory with the three required files; (2)~add its rows, including N/A rows, to the algorithm compatibility tables; and (3)~write per-backend implementation files only for topics where that backend needs a different technique. The library does not cover writing new kernels. For CUDA, Triton, and CUTLASS it points to separate GPU-kernel skills. For Trainium it bundles AWS's NKI skills, pinned to a specific upstream version.

\section{\texorpdfstring{Issue-Tracker MCP Tools}{Issue-Tracker MCP Tools}}
\label{app:mcp-tools}

The issue-tracker policy (\S\ref{sec:design:agent}) exposes its backlog to inner-loop agents through a small stdio MCP server. The backlog is a single JSON file of issues. Each issue has an integer id, a type (\texttt{bug}, \texttt{feature}, or \texttt{perf}), a status (\texttt{open}, \texttt{in\_progress}, \texttt{closed}, or \texttt{blocked}), a one-line title, a Markdown description, the creating agent and round, an attempt counter, and an append-only history of state transitions. The server re-reads the file on every call, so an agent always sees issues filed or updated earlier in the same round. Table~\ref{tab:mcp-tools} lists the four tools.

\begin{table}[h]
\centering
\footnotesize
\caption{Tools exposed by the issue-tracker MCP server. All tools return plain text for the agent to read. An issue's one-line summary has the form \texttt{\#id [type] [status] title}.}
\label{tab:mcp-tools}
\begin{tabular}{@{}lp{3.6cm}p{6.4cm}@{}}
\toprule
Tool & Arguments & Behavior \\
\midrule
\texttt{list\_issues} & \texttt{status} (optional): \texttt{open}, \texttt{in\_progress}, \texttt{closed}, or \texttt{blocked} & One-line summary of every issue, optionally filtered by status. \\
\texttt{get\_issue} & \texttt{issue\_id}: integer & Full issue: type, status, creator and round, attempt count, title, and description. \\
\texttt{search\_issues} & \texttt{query}: comma-separated keywords & One-line summaries of issues whose title or description contains every keyword (case-insensitive). Used to check for duplicates before filing. \\
\texttt{create\_issue} & \texttt{type}: \texttt{bug}, \texttt{feature}, or \texttt{perf}; \texttt{title}: one line; \texttt{description}: Markdown & Files a new \texttt{open} issue if the caller's creation policy allows it; otherwise returns an error explaining why. \\
\bottomrule
\end{tabular}

\end{table}

\paragraph{Creation policy.} The three read tools are unrestricted, but \texttt{create\_issue} is checked against a per-agent policy fixed when the server is launched for that agent's turn. The policy names the creator (recorded on each issue it files), the current round, the issue types the creator may file, and an optional cap on how many issues it may file in that round. A call that violates the policy writes nothing and returns an error the agent can act on: an invalid type lists the valid ones, a disallowed type lists the types this agent may file, and a reached cap tells the agent to deduplicate with \texttt{search\_issues} and prioritize its most impactful issues. The server can also run in read-only mode, in which \texttt{create\_issue} is not registered at all.

\paragraph{Per-role configuration.} Each role that files issues receives its own instance of the server with a policy suited to that role. The Accuracy Judge may file at most one issue per round, and only of type \texttt{bug}; this lets it report a defect unrelated to the issue under review without turning correctness review into open-ended planning. The Performance Evaluator may file \texttt{bug}, \texttt{feature}, and \texttt{perf} issues, up to a configurable per-round cap, and it is the main source of new optimization directions in the backlog. Because the schema is enforced by the server rather than stated only in the prompt, a malformed or over-budget request fails at the tool boundary instead of silently entering the backlog.

\section{Existing Assets and Licenses}
\label{app:licenses}

Table~\ref{tab:assets-licenses} lists the third-party models, datasets, frameworks, and coding-agent harnesses used in this paper, together with their licenses. All assets are used in accordance with their published terms.

\begin{table}[h]
\centering
\footnotesize
\caption{Existing assets used in this paper, with versions, licenses, and source URLs. Citations point to the paper or release we used; see \S\ref{sec:eval} and \S\ref{app:eval-scenarios} for how each asset enters the evaluation.}
\label{tab:assets-licenses}
\setlength{\tabcolsep}{4pt}
\resizebox{\linewidth}{!}{%
\begin{tabular}{@{}lllll@{}}
\toprule
Asset & Type & Version & License & URL \\
\midrule
Llama-3.1-8B-Instruct \cite{grattafiori2024llama} & Model & n/a & Llama 3.1 Community License & \url{https://huggingface.co/meta-llama/Llama-3.1-8B-Instruct} \\
Llama-3.2-1B-Instruct (4-bit, MLX) \cite{grattafiori2024llama} & Model & n/a & Llama 3.2 Community License & \url{https://huggingface.co/mlx-community/Llama-3.2-1B-Instruct-4bit} \\
Qwen3-32B \cite{yang2025qwen3} & Model & n/a & Apache 2.0 & \url{https://huggingface.co/Qwen/Qwen3-32B} \\
Olmo-Hybrid-7B \cite{merrill2026olmo} & Model & n/a & Apache 2.0 & \url{https://huggingface.co/allenai/Olmo-Hybrid-7B} \\
Moonshine Streaming medium \cite{kudlur2026moonshine} & Model & n/a & MIT & \url{https://huggingface.co/UsefulSensors/moonshine-streaming-medium} \\
Show-o2 1.5B-HQ \cite{showo2} & Model & n/a & Apache 2.0 & \url{https://github.com/showlab/show-o} \\
Qwen3.5-397B-A17B-MXFP4 & Model & n/a & Apache 2.0 & \url{https://huggingface.co/amd/Qwen3.5-397B-A17B-MXFP4} \\
\midrule
CodeEditorBench \cite{guo2024codeeditorbench} & Dataset & n/a & Apache 2.0 & \url{https://github.com/CodeEditorBench/CodeEditorBench} \\
JSONSchemaBench \cite{jsonschemabench} & Dataset & n/a & No license specified & \url{https://github.com/guidance-ai/jsonschemabench} \\
\midrule
vLLM \cite{pagedattention} & Framework & v0.19.1 & Apache 2.0 & \url{https://github.com/vllm-project/vllm} \\
SGLang \cite{sglang} & Framework & v0.5.18 & Apache 2.0 & \url{https://github.com/sgl-project/sglang} \\
TensorRT-LLM \cite{tensorrt-llm} & Framework & v1.2.1 & Apache 2.0 & \url{https://github.com/NVIDIA/TensorRT-LLM} \\
HuggingFace Transformers \cite{wolf2019huggingface} & Framework & v5.5.2 & Apache 2.0 & \url{https://github.com/huggingface/transformers} \\
MLX / mlx\_lm \cite{mlx} & Framework & v0.31.2 & MIT & \url{https://github.com/ml-explore/mlx} \\
PyTorch \cite{ansel2024pytorch} & Framework & v2.10 & BSD-3-Clause & \url{https://github.com/pytorch/pytorch} \\
FlashAttention \cite{dao2022flashattention} & Library & fa4-v4.0.0.beta4 & BSD-3-Clause & \url{https://github.com/Dao-AILab/flash-attention} \\
FlashInfer \cite{flashinfer} & Library & v0.6.6 & Apache 2.0 & \url{https://github.com/flashinfer-ai/flashinfer} \\
XGrammar \cite{li2026xgrammar2efficientdynamicstructured} & Library & v0.2.0 & Apache 2.0 & \url{https://github.com/mlc-ai/xgrammar} \\
\midrule
Codex CLI \cite{openai-codex-cli} & Coding-agent harness & v0.125 & Apache 2.0 & \url{https://github.com/openai/codex} \\
Claude Code \cite{anthropic-claude-code} & Coding-agent harness & v2.1.122 & Anthropic ToS (proprietary) & \url{https://www.anthropic.com/claude-code} \\
DeepAgents \cite{langchain-deepagents} & Coding-agent harness & v0.4.11 & MIT & \url{https://github.com/langchain-ai/deepagents} \\
\bottomrule
\end{tabular}%
}
\end{table}

\section{Extended Related Work}
\label{app:related}

This appendix expands the discussion in \S\ref{sec:related}.

Using agents to optimize performance has attracted substantial attention, with recent benchmarks measuring this ability across kernels, numerical routines, repositories, and performance bugs~\citep{kernelbench,press2025algotune,he2025sweperf,ma2025swefficiency,sehgal2025formulacode,garg2025perfbench,shetty2025gso}.
Agentic optimization systems organize around a few search paradigms, none of which have been applied to greenfield end-to-end system implementation and optimization.
\emph{Evolutionary search} maintains a population of agent-generated candidates and selects among them by measured performance. The score numerically encodes the optimization goal, and selection within the population carries that goal forward without summarization, sidestepping the drift that compaction-based handoffs incur.
AlphaEvolve~\citep{alphaevolve}, OpenEvolve~\citep{openevolve}, and SkyDiscover~\citep{skydiscover2026} provide general outer-loop frameworks, and KernelFoundry~\citep{kernelfoundry} and EvoEngineer~\citep{evoengineer} apply this style to GPU kernel generation under paired correctness/performance gates.
As implemented today, these frameworks evolve only small components, e.g., user-marked code regions inside an otherwise-fixed file. A scalar score is sufficient at that scope but cannot encode much of what an end-to-end system needs, e.g., prerequisite dependencies between optimizations (one technique often requires another to be in place first) or internal-bottleneck information that drives the next step, since which component bottlenecks is itself shaped by the agent's prior design choices.
\emph{Multi-agent iteration} replaces the population with agents that hypothesize, experiment, and refine designs across rounds, carrying richer reasoning forward to drive the next decision: Glia~\citep{glia} and Engram~\citep{engram} use this approach to tune systems policies and heuristics, enabling gains on LLM-serving routing and autoscaling, among other tasks.
This reasoning, however, lives within a single context window. Glia's multi-context variant runs independent instances in parallel rather than passing strategic state forward.
A third approach simplifies the loop further: \emph{autoresearch}~\citep{karpathy-autoresearch} puts one long-running agent in charge of the entire search, tracking candidate ideas across git branches, but is prone to drift within the agent's single context.
Adjacent agentic-synthesis work targets similarly bounded scopes. For example, SchedCP~\citep{schedcp} generates Linux scheduling policies without modifying the kernel via LLM-driven techniques.
Across these approaches, the agent's output is a bounded policy, heuristic, or module within a larger system.
\sys is, to our knowledge, the first agentic system under any of these paradigms to do the multi-file coding work needed to design a system itself, creating bespoke LLM serving systems with multiple interconnected internal components.

\sys sits within a broader literature on agents performing long-horizon tasks~\citep{measuring-ai-ability-to-complete-long-tasks,thai2026sweevo,deng2026swebench,anthropic-effective-harnesses,huntley-ralph-loop,anderson2025self}.
The standard recourse when a task exceeds a context window is compaction~\citep{anthropic-context-engineering,kangwook-codex-compaction}, where a session distills its state into a handoff to a fresh successor. Lossy summarization causes drift in both performance and correctness over many rounds, and across optimization sessions an agent must additionally remember which bottleneck to target next, which directions have been tried and discarded, and which platform quirks have surfaced.
\sys is inspired by industrial prototypes for long-horizon coding agents from Cursor and Anthropic, which introduce explicit task abstractions, shared repository state, and custom agent loops so fresh coding-agent sessions can execute small units of work within multi-week autonomous projects~\citep{cursor-scaling-agents,anthropic-c-compiler}. These prototypes show that agent harnesses can build end-to-end systems from scratch, but they stop short of optimizing them.
Git primitives such as commits and branches have also been used to manage agent context and explore distinct strategies across sessions~\citep{wu2026gitcontextcontroller}.
\sys exposes a version-controlled repository interface that allows flexible outer-loop strategies, including the issue-driven loop used in our evaluation.
\sys couples each task with domain-specific correctness and performance gates, so progress is tracked over validated system designs rather than unconstrained repository edits.

\section{\texorpdfstring{Stability Across Independent Runs}{Stability Across Independent Runs}}
\label{app:stability}

Since agentic optimization is stochastic, the same configuration can take different paths through the design space. To quantify this effect, we run \sys three times with the identical configuration for Scenarios D, E, and F.

\paragraph{Scenario D: hybrid-architecture prompt caching.}
We repeated Scenario~D (\S\ref{sec:eval-olmo-hybrid}) three times with an identical configuration and compared the runs at the end of round~5, the last round all three completed. All runs implement Olmo-Hybrid-7B from scratch for the same 20-request, 32k-shared-prefix workload on an L4 GPU, starting from the same repository commit. Table~\ref{tab:stability-olmo} reports the best throughput measured by the trusted benchmark in each run through round~5.

\begin{table}[h]
\centering
\footnotesize
\caption{Three independent Scenario~D runs with an identical configuration, measured at the end of round~5.}
\label{tab:stability-olmo}
\begin{tabular}{lrrrr}
\toprule
 & Run 1 & Run 2 & Run 3 & Mean $\pm$ s.d. \\
\midrule
Best throughput (tok/s) & 123.8 & 108.9 & 151.8 & $128.2 \pm 21.8$ \\
Speedup over vLLM & $5.43\times$ & $4.78\times$ & $6.66\times$ & $(5.62 \pm 0.95)\times$ \\
Time to best candidate (h) & 1.3 & 3.7 & 6.5 & $3.9 \pm 2.6$ \\
Wall-clock time to end of round~5 (h) & 14.7 & 4.3 & 8.8 & $9.3 \pm 5.2$ \\
Agent active time (h) & 1.6 & 2.0 & 2.8 & $2.1 \pm 0.6$ \\
Agent turns & 30 & 16 & 24 & $23 \pm 7$ \\
Input tokens (M) & 33.8 & 31.5 & 37.3 & $34.2 \pm 3.0$ \\
\quad served from the prompt cache & 96.3\% & 97.6\% & 97.0\% & $97.0 \pm 0.7$\% \\
Output tokens (k) & 161.0 & 161.3 & 248.6 & $190.3 \pm 50.5$ \\
\quad of which reasoning & 61.2 & 77.4 & 134.4 & $91.0 \pm 38.5$ \\
\bottomrule
\end{tabular}

\end{table}

All three runs pass the accuracy gate and exceed vLLM by $4.8$--$6.7\times$ within five rounds. Although the final performance numbers are similar, each follows a different trajectory. Run~1 found its best candidate in round~1 (1.3\,h) and did not improve on it in subsequent rounds. Run~2 recorded its only official round score in round~3, while Run~3 improved steadily and reached the best performance at round~3.

\paragraph{Scenario E: streaming ASR.}
We repeated Scenario~E (\S\ref{sec:eval-moonshine}) three times with an identical configuration and compared the runs at the end of round~5. All runs implement Moonshine Streaming medium from scratch for the same workload (32 concurrent streams, a 2-second audio chunk every 2 seconds) on an L4 GPU. Table~\ref{tab:stability-moonshine} reports the best TTFT measured in each run through round~5 and the corresponding speedup over the vLLM-plugin baseline from \S\ref{sec:eval-moonshine}.

\begin{table}[h]
\centering
\footnotesize
\caption{Three independent Scenario~E runs with an identical configuration, measured at the end of round~5.}
\label{tab:stability-moonshine}
\begin{tabular}{lrrrr}
\toprule
 & Run 1 & Run 2 & Run 3 & Mean $\pm$ s.d. \\
\midrule
Best TTFT p50 (ms) & 117.6 & 186.2 & 220.5 & $174.7 \pm 52.4$ \\
Speedup over vLLM plugin & $4.59\times$ & $2.90\times$ & $2.45\times$ & $(3.31 \pm 1.13)\times$ \\
Time to best candidate (h) & 5.1 & 2.4 & 5.2 & $4.2 \pm 1.6$ \\
Wall-clock time to end of round~5 (h) & 5.1 & 6.6 & 5.2 & $5.6 \pm 0.9$ \\
Agent active time (h) & 1.4 & 2.6 & 1.8 & $1.9 \pm 0.6$ \\
Agent turns & 24 & 28 & 24 & $25 \pm 2$ \\
Input tokens (M) & 24.3 & 39.7 & 25.6 & $29.9 \pm 8.6$ \\
\quad served from the prompt cache & 95.6\% & 96.7\% & 96.2\% & $96.2 \pm 0.5$\% \\
Output tokens (k) & 134.7 & 226.6 & 156.1 & $172.5 \pm 48.1$ \\
\quad of which reasoning & 46.4 & 95.9 & 50.6 & $64.3 \pm 27.4$ \\
\bottomrule
\end{tabular}

\end{table}

All three runs pass the accuracy gate and beat the vLLM plugin by $2.5$--$4.6\times$ within five rounds. The spread is wider than in Scenario~D, and the trajectories differ. Run~1 improved in its last round, from 165.7 to 117.6\,ms. Run~2 reached a best TTFT of 186.2\,ms in round~3. Run~3 scored 263--304\,ms in rounds~1--4 and improved to 220.5\,ms in round~5.

\paragraph{Scenario F: local constrained decoding.}
We repeated Scenario~F (\S\ref{sec:eval-spec-cons-mbp}) three times with an identical configuration. All runs optimize Llama-3.1-8B-Instruct for single-stream JSON generation under constrained decoding on JSONSchemaBench on the same MacBook Pro, with each run fixed to a 3-hour wall-clock budget. Table~\ref{tab:stability-json} compares the runs at the end of their 3-hour budgets. Latency is the best trusted p50 benchmark result, reported against the mlx\_lm baseline.

\begin{table}[h]
\centering
\footnotesize
\caption{Three independent Scenario~F runs with an identical configuration.}
\label{tab:stability-json}
\begin{tabular}{lrrrr}
\toprule
 & Run 1 & Run 2 & Run 3 & Mean $\pm$ s.d. \\
\midrule
Best p50 latency (ms) & 1525 & 1652 & 1488 & $1555 \pm 86$ \\
Speedup over mlx\_lm & $2.25\times$ & $2.08\times$ & $2.30\times$ & $(2.21 \pm 0.12)\times$ \\
Time to best candidate (h) & 2.1 & 2.8 & 2.3 & $2.4 \pm 0.4$ \\
Agent active time (h) & 1.37 & 1.63 & 1.54 & $1.51 \pm 0.13$ \\
Agent turns & 104 & 128 & 108 & $113 \pm 13$ \\
Input tokens (M) & 75.1 & 79.3 & 77.2 & $77.2 \pm 2.1$ \\
\quad served from the prompt cache & 98.0\% & 98.5\% & 97.7\% & $98.1 \pm 0.4$\% \\
Output tokens (k) & 488.5 & 513.8 & 523.7 & $508.7 \pm 18.2$ \\
\bottomrule
\end{tabular}

\end{table}

All three runs reach a $2.08$--$2.30\times$ speedup over mlx\_lm within the 3-hour budget. They also converge on the same core optimizations: emitting grammar-forced tokens without running the model, KV caching, speculative decoding under the grammar constraint, and tuning the draft length. Beyond these, each run tried different optimizations. Run~1 distilled the 1B draft model on the 8B target's own outputs to raise draft accuracy, but concluded that the resulting 3.8\% reduction in end-to-end latency was not statistically significant. Run~2 pinned CPU-side threads to avoid interference with other CPU processes. Run~3 pipelined the speculative rounds so that CPU and GPU work overlap, but this was 1.6--8\% slower and was abandoned.

\section{\texorpdfstring{Behavior Under Modified Traffic}{Behavior Under Modified Traffic}}
\label{app:traffic}

\sys specializes each system to its target workload, so a natural question is how the resulting systems behave when the traffic changes. We take the final systems from Scenarios~A and C--G (\S\ref{sec:eval-standard} and \S\ref{sec:eval-qwen-code-edit}--\S\ref{sec:eval-show-o2-mac}) without modification and run them on a set of perturbed workloads. Each perturbation changes one property of the workload, such as concurrency, request count, sequence lengths, arrival pattern, or input distribution. Note that for Scenario~B, the goodput number we reported already comes from traffic the agents did not optimize against, as performance evaluation uses a 20\% held-out set of the workload (Appendix~\ref{app:eval-scenarios}).

\paragraph{Scenario A: standard LLM serving.}
The original workload draws prompts from a fixed pool and caps outputs at 128 tokens. We replace it with two fixed input/output shapes at several concurrency levels: a long-input shape (7{,}168 input and 128 output tokens), where prefill dominates, and a decode-heavy shape (256 input and 1{,}024 output tokens), where decoding dominates. The final Scenario~A system uses speculative decoding, so we compare it against vLLM both with and without speculative decoding (n-gram match length 1-4, tuned to work best for the synthetic workload). The headline results in \S\ref{sec:eval-standard} compare against vLLM without speculative decoding, because that workload is throughput-oriented and speculative decoding does not help for aggregated throughput. By contrast, workloads here include decode-heavy and low-concurrency ones, so we also enable it in vLLM for a fairer comparison. Table~\ref{tab:traffic-llama} reports output token throughput.

\begin{table}[h]
\centering
\footnotesize
\caption{Scenario~A under modified traffic: output token throughput (tok/s) on Llama-3.1-8B-Instruct (H100). ``Speedup'' is \sys{}'s throughput over the better of the two vLLM configurations.}
\label{tab:traffic-llama}
\begin{tabular}{@{}rrrrr@{}}
\toprule
Concurrency & vLLM (no spec.) & vLLM (spec.) & \sys{} & Speedup \\
\midrule
\multicolumn{5}{@{}l}{\emph{Long input (7{,}168 in / 128 out)}} \\
4 & 311 & 498 & 482 & $0.97\times$ \\
16 & 409 & 542 & 530 & $0.98\times$ \\
32 & 484 & 539 & 534 & $0.99\times$ \\
64 & 540 & 541 & 532 & $0.98\times$ \\
128 & 538 & 531 & 531 & $0.99\times$ \\
\midrule
\multicolumn{5}{@{}l}{\emph{Decode heavy (256 in / 1{,}024 out)}} \\
4 & 613 & 2,122 & 2,184 & $1.03\times$ \\
16 & 2,254 & 7,090 & 6,821 & $0.96\times$ \\
32 & 4,090 & 9,648 & 9,839 & $1.02\times$ \\
64 & 6,972 & 13,011 & 12,968 & $1.00\times$ \\
128 & 10,444 & 13,591 & 13,196 & $0.97\times$ \\
\bottomrule
\end{tabular}

\end{table}

Across both shapes, \sys{} stays within 4\% of vLLM with speculative decoding ($0.96$--$1.03\times$) at every concurrency, so its performance carries over to input/output shapes it was not optimized for. Its advantage over vLLM without speculative decoding comes from speculative decoding and shrinks as batches grow.

\paragraph{Scenario C: predicted-output code editing.} We compare against two vLLM configurations: vanilla decoding and EAGLE3 speculative decoding with 7 speculative tokens. \begin{itemize}[leftmargin=*,itemsep=0pt,topsep=2pt]
\item \textbf{Original:} the optimization workload. Python edits from CodeEditorBench sent one at a time, with greedy decoding, up to 512 output tokens, and the pre-edit (buggy) program as the prediction.
\item \textbf{C++ and Java:} CodeEditorBench edits in C++ or Java instead of Python.
\item \textbf{$4\times$ longer files:} four pre-edit programs concatenated into one file per request, with up to 2{,}048 output tokens.
\item \textbf{Concurrency 4:} four requests in flight at once instead of one.
\item \textbf{No prediction:} requests omit the \texttt{prediction} field.
\item \textbf{Wrong prediction:} each request carries another sample's pre-edit program as its prediction, so the prediction rarely matches the output.
\end{itemize}
Table~\ref{tab:traffic-code-edit} reports median per-request throughput.

\begin{table}[h]
\centering
\footnotesize
\caption{Scenario~C under modified traffic: median per-request decode throughput (tok/s) on Qwen3-32B (H100). vLLM ignores the \texttt{prediction} field, so its ``No prediction'' and ``Wrong prediction'' rows repeat the original workload.}
\label{tab:traffic-code-edit}
\setlength{\tabcolsep}{4pt}
\begin{tabular}{@{}lrrrr@{}}
\toprule
Workload & vLLM & vLLM EAGLE3 & \sys & \sys{} / EAGLE3 \\
\midrule
Original & 40.8 & 122.7 & 284.3 & $2.32\times$ \\
C++ & 40.9 & 122.5 & 396.1 & $3.24\times$ \\
Java & 41.0 & 107.2 & 340.8 & $3.18\times$ \\
$4\times$ longer files & 41.0 & 86.6 & 333.2 & $3.85\times$ \\
Concurrency 4 & 40.3 & 100.6 & 71.2 & $0.71\times$ \\
No prediction & 40.8 & 122.7 & 38.5 & $0.31\times$ \\
Wrong prediction & 40.8 & 122.7 & 38.4 & $0.31\times$ \\
\bottomrule
\end{tabular}

\end{table}

With a correct prediction at concurrency~1, the system keeps a $7$--$10\times$ advantage over vanilla vLLM and a $2.3$--$3.9\times$ advantage over vLLM EAGLE3 across languages and longer files, so the predicted-output verifier is not tied to the benchmark's Python files. Two limitations appear outside the original workload. First, the system serializes requests: at concurrency~4 its per-request throughput drops from 284.3 to 71.2~tok/s, below vLLM EAGLE3's 100.6~tok/s. Second, without a correct prediction it gains nothing over vanilla vLLM (38.5 vs.\ 40.8~tok/s).

\paragraph{Scenario D: hybrid-architecture prompt caching.}
The original workload sends 20 concurrent requests, each with a 32{,}768-token shared prefix, a 128-token unique suffix, and 128 output tokens. \begin{itemize}[leftmargin=*,itemsep=0pt,topsep=2pt]
\item \textbf{16{,}384-token shared prefix:} a shorter shared prefix.
\item \textbf{One arrival per second:} request $i$ starts $i$ seconds after the first instead of all requests arriving at once.
\item \textbf{256 output tokens:} longer outputs (256 instead of 128 tokens).
\item \textbf{8 requests:} fewer requests than the 20 of the original workload.
\end{itemize}
Table~\ref{tab:traffic-olmo} reports aggregate throughput and the number of requests that completed.

\begin{table}[h]
\centering
\footnotesize
\caption{Scenario~D under modified traffic: aggregate throughput (tok/s) on Olmo-Hybrid-7B (L4), with completed requests in parentheses. ``Fails'' means some requests did not complete.}
\label{tab:traffic-olmo}
\begin{tabular}{@{}lrrrl@{}}
\toprule
Workload & \sys & vLLM & \sys{} / vLLM & \sys failure mode \\
\midrule
Original                                & 69.9 (20/20) & 22.8 (20/20)  & $3.06\times$ & -- \\
16{,}384-token shared prefix            & 87.2 (20/20) & 133.6 (20/20) & $0.65\times$ & -- \\
One arrival per second                  & 53.0 (20/20) & 21.4 (20/20)  & $2.47\times$ & -- \\
256 output tokens                       & 0.0 (0/20)   & 37.4 (20/20)  & fails & KV pool full \\
8 requests                              & 0.0 (0/8)    & 57.6 (8/8)    & fails & hang \\
\bottomrule
\end{tabular}

\end{table}

The system keeps its advantage when requests arrive over time. It falls behind vLLM with a 16{,}384-token shared prefix, a length it was not designed for, since it was built for the 32{,}768-token prefix of the target workload. It is also hard-wired to the benchmark's shape of 20 requests with 128 output tokens, and changing either one makes it fail, by exhausting its KV pool or waiting indefinitely for a full batch of 20 requests.

\paragraph{Scenario E: streaming ASR.}
The configuration the optimization run measured uses 32 concurrent streams sending 0.5-second audio chunks. \begin{itemize}[leftmargin=*,itemsep=0pt,topsep=2pt]
\item \textbf{1\,s and 2\,s chunks:} each stream sends 1- or 2-second audio chunks instead of 0.5-second chunks, at real-time pace.
\item \textbf{1, 8, and 64 streams:} fewer or more concurrent streams than the original 32.
\item \textbf{LibriSpeech:} LibriSpeech test-clean utterances instead of the benchmark's audio clips.
\end{itemize}
Table~\ref{tab:traffic-moonshine} reports the results.

\begin{table}[h]
\centering
\footnotesize
\caption{Scenario~E under modified traffic on Moonshine Streaming (L4). TTFT is the p50 latency (ms) from the first chunk to the first computed partial transcript.}
\label{tab:traffic-moonshine}
\setlength{\tabcolsep}{4pt}
\begin{tabular}{@{}lrrr@{}}
\toprule
 & \multicolumn{3}{c}{TTFT p50 (ms)} \\
\cmidrule(l){2-4}
Workload & \sys & vLLM & Speedup \\
\midrule
2\,s chunks                         & 708 & 5786 & $8.17\times$ \\
1\,s chunks                         & 731 & 773  & $1.06\times$ \\
1 stream                            & 171 & 259  & $1.51\times$ \\
8 streams                           & 221 & 794  & $3.60\times$ \\
64 streams                          & 777 & 938  & $1.21\times$ \\
LibriSpeech                         & 518 & 1597 & $3.08\times$ \\
\bottomrule
\end{tabular}

\end{table}

The system is faster than vLLM on every perturbed workload ($1.06$--$8.17\times$), across chunk sizes, concurrency levels, and LibriSpeech audio. Among concurrency levels, the advantage is largest at 1 and 8 streams ($1.51\times$ and $3.60\times$), and across chunk sizes it is smallest with 1-second chunks ($1.06\times$).

\paragraph{Scenario F: local constrained decoding.}
We run the final system on JSONSchemaBench's held-out set, which was not used during optimization. Its schemas and prompts are slightly longer than those of the benchmark set (17\% longer in p90), so both mlx\_lm and \sys{} have higher latency on it. The mean latency falls from 3{,}563\,ms with mlx\_lm to 1{,}576\,ms with \sys{}, which is a $2.26\times$ speedup, similar to the $2.6\times$ on the benchmark set (\S\ref{sec:eval-spec-cons-mbp}).

\paragraph{Scenario G: local image generation.}
We run the final MacBook system with two changes to the original workload of one fixed prompt at 20 sampler steps. \begin{itemize}[leftmargin=*,itemsep=0pt,topsep=2pt]
\item \textbf{Unseen prompts:} 60 VBench~\cite{huang2023vbench} prompts not used during optimization, 10 from each of six suites, at $512{\times}512$ instead of $432{\times}432$, with 20 sampler steps and guidance scale 5.0. These prompts test whether that approximation lowers output quality on prompts not used during optimization. The final system and the PyTorch-MPS baseline generate each image with the same seed.
\item \textbf{50 denoising steps:} 50 sampler steps instead of the 20 the system was optimized for.
\end{itemize}
On the unseen prompts, the system takes 20.9\,s per image against 91.9\,s for the baseline, a $4.4\times$ speedup, and every request runs on the system's fast path. Its images match their prompts as well as the baseline's images do on both SigLIP and CLIP-L image-text scores. On VBench's image metrics, the system's aesthetic quality is 2\% lower. Its imaging quality is 3.5\% higher because its images are sharper and higher in contrast. With 50 sampler steps, the system's latency grows linearly with the number of steps, as expected, and its mean latency is $6.26\times$ lower than the PyTorch-MPS baseline's, in line with the $6.27\times$ speedup at 20 steps (\S\ref{sec:eval-show-o2-mac}).

\end{document}